%% file: samplepaper.tex
\documentclass[runningheads]{llncs}
\usepackage[T1]{fontenc}
\usepackage{orcidlink}
\usepackage{makecell}
\usepackage[export]{adjustbox}
\usepackage{xcolor}
\usepackage{tikz}
\usetikzlibrary{positioning,arrows.meta}
\usepackage{booktabs}
\usepackage{multirow}

\usepackage{subcaption}
\usepackage{amsmath}
\usepackage{amssymb}
\usepackage{pifont}
\newcommand{\cmark}{\ding{51}}%
\newcommand{\xmark}{\ding{55}}%
\newcommand{\bull}{\textsc{BullingerDB}}
\usepackage[utf8]{inputenc}
\usepackage{pgfplots}
\usepackage{adjustbox}
\DeclareUnicodeCharacter{2212}{−}
\usepgfplotslibrary{groupplots,dateplot}
\usetikzlibrary{patterns,shapes.arrows}
\pgfplotsset{compat=newest}

\usepackage{graphicx}
\usepackage[nolist]{acronym}
\usepackage{hyperref}

\begin{document}

\newacro{AP}[AP]{Average Precision}
\newacro{cnn}[CNN]{Convolutional Neural Network}
\newacro{dl}[DL]{Deep Learning}
\newacro{dnn}[DNN]{Deep Neural Network}
\newacro{HTR}[HTR]{Handwritten Text Recognition}
\newacro{knn}[kNN]{k-nearest neighbors}
\newacro{mAP}[mAP]{mean Average Precision}
\newacro{mlp}[MLP]{multilayer perceptron}
\newacro{par}[P@$r$]{Precision at $r$}
\newacro{sift}[SIFT]{Scale-Invariant Feature Transform}
\newacro{vit}[ViT]{Vision Transformer}
\newacro{vlad}[VLAD]{Vector of Locally Aggregated Descriptors}
\newacro{wi}[WI]{Writer Identification}
\newacro{WR}[WR]{Writer Retrieval}
\newacro{nDCG}{Normalized Discounted Cumulative Gain}
\newacro{CER}{Character Error Rate}
\newacro{WER}{Word Error Rate}
\newacro{CTC}{Connectionist Temporal Classification}

\title{\bull: A Dataset for Handwritten Text Recognition and Writer Retrieval}
\titlerunning{\bull: A Benchmark for HTR and WR}
%
\author{Marco Peer\inst{1}\orcidlink{0000-0001-6843-0830} \and
Anna Scius-Bertrand\inst{1,2} \and 
Patricia Scheurer\inst{3}\orcidlink{0009-0007-4146-955X} \and
Andreas Fischer\inst{1}\orcidlink{0000-0003-0069-3436}}
\authorrunning{M. Peer et al.}
\institute{%
AIBEX Group, University of Fribourg, Switzerland
\and
iCoSys Institute, University of Applied Sciences and Arts Western Switzerland
\and
Department of Computational Linguistics, University of Zurich, Switzerland
}
\maketitle              
\begin{abstract}
We present \bull, a large-scale benchmark dataset for historical document analysis based on the correspondence of Heinrich Bullinger (1504–1575). The corpus comprises 20,898 pages and 499,222 text lines written by 796 writers over six decades, featuring stylistic variation, multilingual content (mostly Latin and Early New High German) as well as meta-information such as writer identity and time.
We evaluate \bull \ on text recognition and writer retrieval. TrOCR, the best performing model, achieves a CER of 9.1\%. For writer retrieval, we introduce a temporal nDCG metric to assess time-aware retrieval. While temporally coherent retrieval is achievable, mAP (78.3\%) scores indicate challenges due to long-term stylistic variation.
With \bull, we aim to establish a new benchmark for multilingual historical text recognition and temporally-aware writer analysis.
\end{abstract}

\input{sections/1_introduction}
\input{sections/2_related_work}

\input{sections/3_dataset}

\input{sections/4_methods}

\input{sections/5_experiments}
\input{sections/6_conclusion}

%
%
%
\bibliographystyle{splncs04}
\bibliography{bib}
\end{document}

%% file: sections/1_introduction.tex
\section{Introduction}

In this paper, we present \bull, a large-scale dataset consisting of the correspondance of Heinrich Bullinger for historical document analysis. It is based on the letters written and received by Heinrich Bullinger (1504-1575). They form a large Reformation-era letter corpus that comprises more than 12,000 letters, including around 2,000 written by Bullinger himself \cite{Mauelshagen2010}. The letters (usually consisting of 2 -- 3 pages) were exchanged with recipients across Europe and are preserved in editions and digital form. Furthermore, they span six decades of the 16th century and are written in multiple languages, with Latin and Early New High German dominating \cite{stroebel2024multilingual}. 

\begin{figure}[h]
    \centering
        \begin{subfigure}[t]{0.66\textwidth}
        \centering
        \includegraphics[
            height=6cm,
            trim=0 0 0 0, 
            clip
        ]{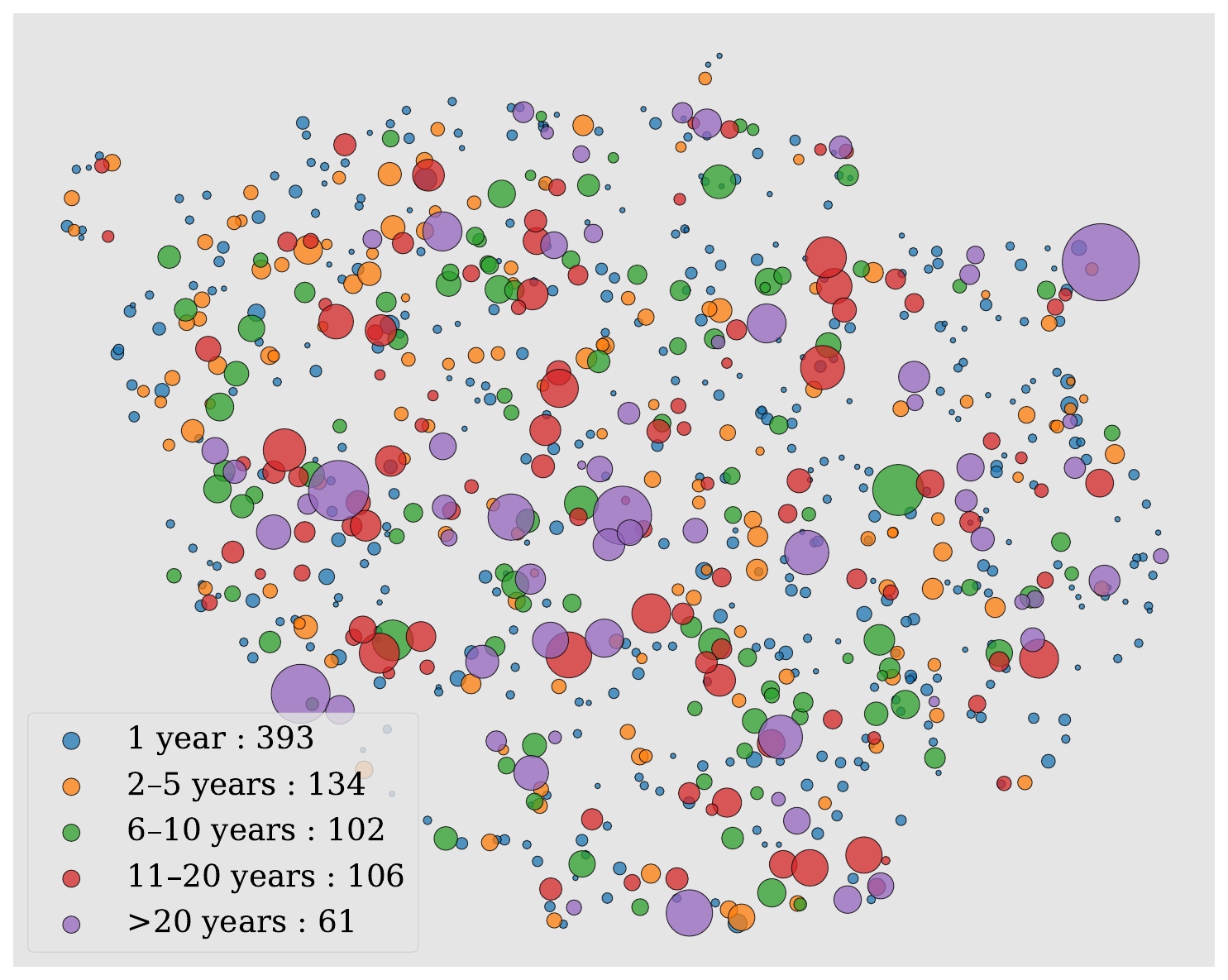}
        \caption{}
        \label{fig:tsne}
    \end{subfigure}
            \hfill
    \begin{subfigure}[t]{0.33\textwidth}
        \centering
        \includegraphics[
            height=5.85cm,
            trim=0 0 0 0,
            clip
        ]{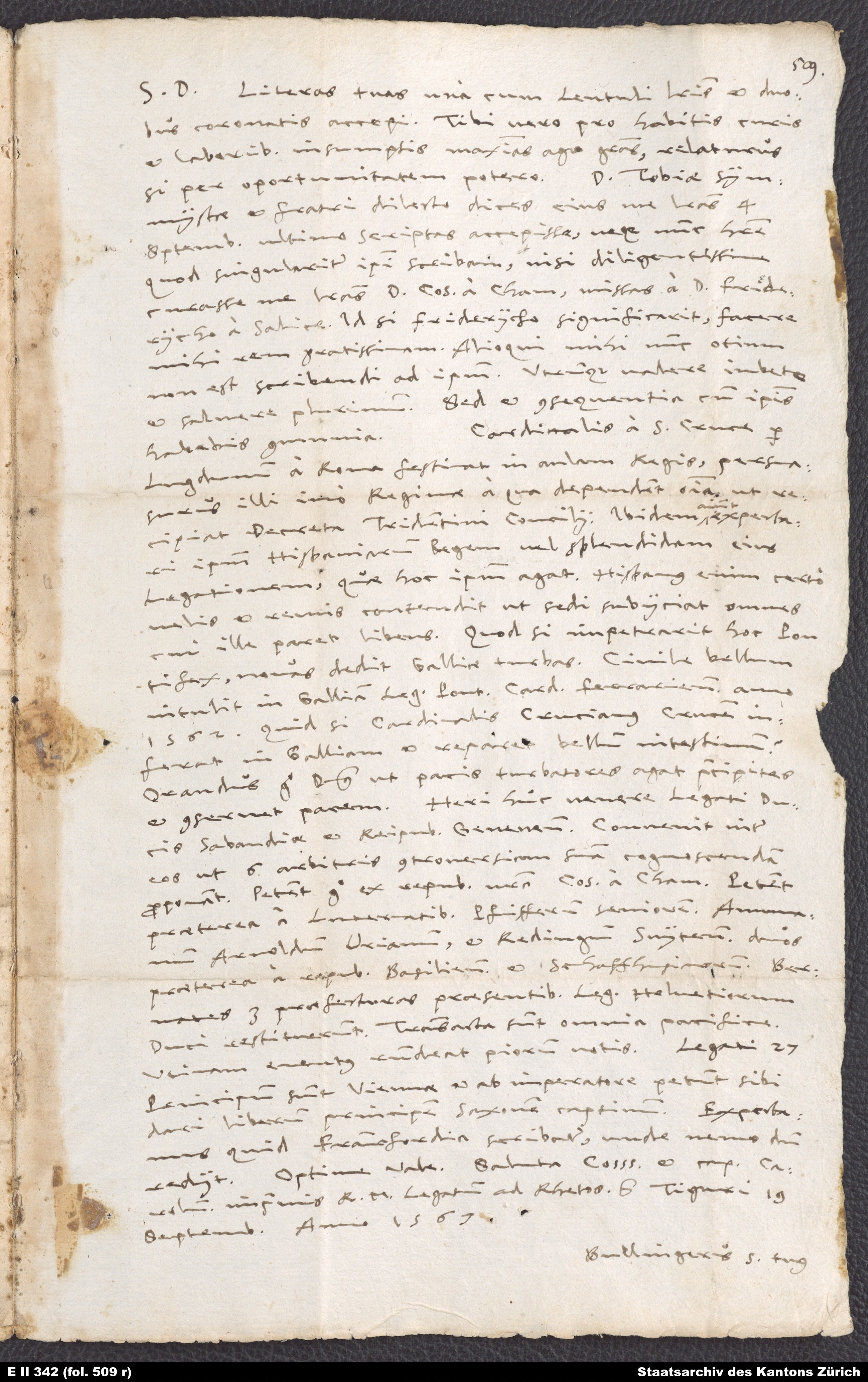}
        \caption{}
        \label{fig:sample}
    \end{subfigure}

    \caption{(a) U-MAP visualization of \bull. Each dot represents a writer with their style embedding, the size corresponds to the number of pages, and the color to the time span the writer was active. (b) Example page written by Heinrich Bullinger.}
    \label{fig:combined}
\end{figure}

From a document analysis perspective, \bull \ constitutes a highly challenging and valuable corpus. Our dataset comprises 20,898 pages of 796 individual writers, spans six decades (1523 -- 1575), and exhibits a large variation in handwriting styles. A first, smaller subset of the letters was introduced to the document analysis community as a benchmark for writer adaptation and historical handwriting recognition by Scius-Bertrand et al. \cite{bullinger-dataset2023} (consisting of 8,000 pages). However, the \href{https://www.bullinger-digital.ch/index.html}{Bullinger Digital} project has established a long-term scholarly digital edition of the correspondence, providing structured metadata and TEI-encoded texts \cite{stroebel2024multilingual}. While the corpus is thus well documented from editorial and NLP perspectives, it currently lacks transcriptions suitable for training and evaluating modern \ac{HTR} systems. With \bull, we aim to close this gap and publish the full collection of Bullinger's correspondance.

We focus on two core tasks in our work to show the impact of \bull. First, \ac{HTR}: The letters are written - among other, less frequent languages, such as Greek - not only in Latin but also in a premodern form of German, with frequent code-switching that may occur between paragraphs or even within a single sentence \cite{volk-etal-2022-nunc}. This multilingual and historically variable language use, combined with variable and difficult handwriting styles, poses challenges even for human experts (a sample is shown in Fig. \ref{fig:sample}). At the same time, the corpus comprises half a million text lines, that enables studies with large-scale experiments and training of deep learning models from scratch.

Secondly, we address the task of \ac{WR}, since \bull \ also provides precise and scholar-validated attributions of 796 writers, along with rich temporal metadata. We show a visualization of the handwriting styles in Fig. \ref{fig:tsne}, that also highlights the timespan in which writers contributed letters, with Bullinger himself having 2,986 pages in our dataset. This time aspect makes \bull\ unique in the field of \ac{WR} and particularly interesting for studying stylistic changes over time (e.g., handwriting evolution with age) as well as enabling training of text-dependent approaches that exploit the availability of transcriptions. Hence, we introduce a new metric - the temporal \ac{nDCG} -  for benchmarking \ac{WR} systems to consider temporal coherence of the retrieval.

Our results show that for \ac{HTR}, the best-performing model, TrOCR - a transformer-based architecture - achieves a \ac{CER} of 9.1\% despite the diverse handwriting styles in the dataset, with performance degrading particularly for shorter lines where contextual information is limited and for some writing styles that are not in the training data. For \ac{WR}, current state-of-the-art approaches are able to produce temporally coherent retrieval on \bull, as reflected by strong nDCG scores; however, the comparatively lower \ac{mAP} values indicate that overall ranking quality is negatively affected by stylistic changes over time, suggesting that long-term stylistic variation remains an issue for robust \ac{WR}, in particular for supervised training.

To summarize, our contributions of the paper are as follows:
\begin{itemize}
    \item We propose \bull, a large scale historical dataset consisting of 20,898 pages contributed by 796 writers. The dataset includes line segmentations, transcriptions and meta information about the writer and date. The dataset is publicly available via \href{https://zenodo.org/records/19728926}{zenodo}.
    \item We provide baseline results and evaluation protocols for the task of 1) \ac{HTR} and 2) \ac{WR} to support future research and benchmarking. For \ac{WR}, we also introduce the \ac{nDCG} metric that measures temporal consistency. 
\end{itemize}

In the remainder of this work, we describe related work (Section \ref{sec:rel_work}), explain the dataset and processing pipeline (Section \ref{sec:dataset}) and present the methods (Section \ref{sec:methods}). Finally, we show the results in Section \ref{sec:results} and provide a conclusion in Section~\ref{sec:conclusion}.\newline


%% file: sections/2_related_work.tex
\section{Related Work}\label{sec:rel_work}

In the following, we explore the task of \ac{HTR} and \ac{WR} for historical document analysis, with a focus on the dataset perspective.

\subsection{Handwritten Text Recognition}

Modern offline \ac{HTR} systems are dominantly formulated as sequence-to-sequence problems at line level and trained with the \ac{CTC} loss \cite{Graves2006}. A common architectural pattern combines CNNs for visual feature extraction with recurrent layers, such as bidirectional LSTMs \cite{Hochreiter1997}, for sequence modeling, followed by a CTC decoding layer. This CNN–RNN–CTC paradigm has been widely adopted for historical manuscripts due to its robustness to variable-length inputs and heterogeneous handwriting styles \cite{Fischer2020-al}, e.g. in \cite{deSousaNeto2020,Retsinas2022,PyLaia}. With the transformer architecture arising in sequence modelling, it has also been applied to \ac{HTR} by formulating the task as autoregressive sequence generation, where TrOCR \cite{trocr}, a transformer encoder–decoder architecture, presents a popular architecture, in particular for finetuning.

While research is focused on studying and and optimizing architectures for \ac{HTR}, the field still lacks a challenging, historical large-scale dataset. With NorhandV3 \cite{norhand}, consisting of about 250k lines, another dataset aimed at medieval Norwegian handwriting is proposed. However, as we show in Table \ref{tab:htr_datasets}, the proposed \bull \ is twice as large than NorhandV3 and more than ten times larger than contemporary \ac{HTR} benchmarks such as IAM \cite{marti_iam-database_2002}.

In the course of the Bullinger Digital project, efforts towards discovering the correspondances of Heinrich Bullinger have been made. Stroebel et al. \cite{Strbel2023} investigate the application of TrOCR, while Scius-Bertrand et al. \cite{bullinger-dataset2023} evaluate HTRFlor \cite{deSousaNeto2020} and PyLaia \cite{PyLaia}. Both evaluate on a subset of the letters, with CER values ranging from 6-10\% usually, depending on the writer setup. Similarly, Spoto et al. work on improving accuracy by using synthetic data with a generative model \cite{Spoto2022}. Besides \ac{HTR}, Volk et al. \cite{volk-etal-2022-nunc} explore the detection of code-switching in the corpus, which mostly happens between German and Latin. Since around 3,100 edited transcriptions done by scholars are available \cite{Strbel2023}, a method for text to image alignment (line-based) is proposed by Peer at al. \cite{Peer2025}. This approach is also used in this work to refine the \ac{HTR} model used to generate the final transcriptions of \bull. Finally, the general workflow of how the Bullinger Digital project digitalized the collection as well as metadata extraction and annotation is described in \cite{stroebel2024multilingual}.

\begin{table*}
    \centering
    \caption{Comparison of \bull \ to existing benchmark datasets.}
    \label{tab:datasets}

    \begin{subtable}[t]{0.38\textwidth}
        \centering
        \caption{HTR}
        \label{tab:htr_datasets}
        \begin{tabular}{lrr}
            \toprule
            Dataset & Lines & Writers \\ 
            \midrule
            GW \cite{lavrenko2004holistic} & 656 & 2 \\
            Parzival \cite{fischer2012lexiconfree} & 4,477 & 3 \\
            NorhandV3 \cite{norhand} & 248,355 & \emph{n/a} \\
            IAM \cite{marti_iam-database_2002} & 13,353 & 657 \\
            \midrule
            \bull & 499,222 & 796 \\
            \bottomrule
        \end{tabular}
    \end{subtable}
    \hfill
    \begin{subtable}[t]{0.6\textwidth}
        \centering
        \caption{WR}
        \label{tab:wr_datasets}
        \begin{tabular}{lllccc}\toprule
            & Pages & Writers & Layout & Text & Date \\ \midrule
            CVL \cite{cvl} & 1,409 & 283 & \cmark & \cmark & \xmark \\
            IAM \cite{marti_iam-database_2002} & 1,539 & 657 & \cmark & \cmark & \xmark \\
            Historical-WI \cite{historical-wi} & 3,600 & 720 & \xmark & \xmark & \xmark \\ 
            HisIR19 \cite{icdar19} & 20,000 & 10,068 & \xmark & \xmark & \xmark \\
            \midrule
            \bull & 20,898 & 796 & \cmark & \cmark & \cmark \\ \bottomrule
        \end{tabular}
    \end{subtable}

\end{table*}

\subsection{Writer Retrieval}
\ac{WR} has received attention in particular in the domain of historical manuscripts, for which unsupervised approaches, e.g., clustering SIFT descriptors and using the assignment as training signal for a CNN (Cl-S) \cite{unsupervised_icdar17}. or self-supervised schemes such as DINO \cite{peer2022}, masked autoencoders \cite{peer2024} and AttMask \cite{raven}, outperform supervised training protocols. The page descriptors are then encoded via VLAD-based codebooks (VLAD \cite{unsupervised_icdar17,raven} or NetVLAD variants \cite{peer_netrvlad,peer2024}).

Regarding the evaluation datasets, an overview is given in Table \ref{tab:wr_datasets}. Historical-WI~\cite{historical-wi} and HisIR19~\cite{icdar19} are the current default datasets. However, they do not provide layout, textual, or temporal metadata. Such information is typically only available for contemporary handwriting datasets (e.g., CVL~\cite{cvl} and IAM~\cite{marti_iam-database_2002}), which were collected in controlled settings. These datasets are considerably smaller and lack temporal variation though. \bull \ is therefore the largest publicly available dataset among related work, comprising 20,898 pages (see Table~\ref{tab:wr_datasets}). In addition to its scale, the dataset proposed provides metadata, including layout information, transcriptions, and document dates.  

Considering the number of writers, HisIR19 includes a larger total number of contributors. Nevertheless, 7,500 of its writers provide only a single sample, resulting in limited intra-writer variability. This makes our dataset proposed particularly well-suited for studying writer-specific characteristics across multiple samples and also adds difficulty to supervised training protocols.

Finally, our dataset presents the unique challenge to study the evolvement of handwriting of a single person over time. This aspect is also conceptually related to the \emph{AnyScript: Long-Term Handwriting Author Identification Challenge} \cite{anyscript2026} (taking place at ICDAR 2026), which addresses temporally separated handwriting comparison in an identification setting. In contrast to AnyScript, \bull \ is temporally dense: all writers are active within the same century, and the documents exhibit consistent layout conventions and content structure. As a consequence, writer-specific characteristics are less influenced by large-scale historical variation in script style or document format. We hypothesize that this setting increases the difficulty of \ac{WR}, as discriminative cues must be derived from individual handwriting traits rather than from broader stylistic or chronological differences.

%% file: sections/3_dataset.tex
\section{Dataset}\label{sec:dataset}

We describe the relevance of \bull \ and explain the processing pipeline in the following.

\subsection{Historical Context of \bull}

The corpus comprises the surviving correspondence of the Swiss reformer Heinrich Bullinger (1504--1575), a central figure of sixteenth-century Protestantism. Content-wise, as one of the largest extant Reformation-era corpora, the letters provide a dense, longitudinal record of intellectual exchange during a formative period of Europe. The correspondence documents the consolidation of reformed Protestantism, the negotiation of emerging confessional identities, and the circulation of theological and political arguments across territorial and linguistic boundaries \cite{HBBW_Reihe}. For historians, the corpus constitutes a main source for reconstructing social and intellectual networks of the Reformation and for studying multilingual written culture in the sixteenth century \cite{Mauelshagen2010}.

\begin{figure}[h]
\centering
\begin{subfigure}[b]{0.6\linewidth}
    \centering
    \includegraphics[width=\linewidth]{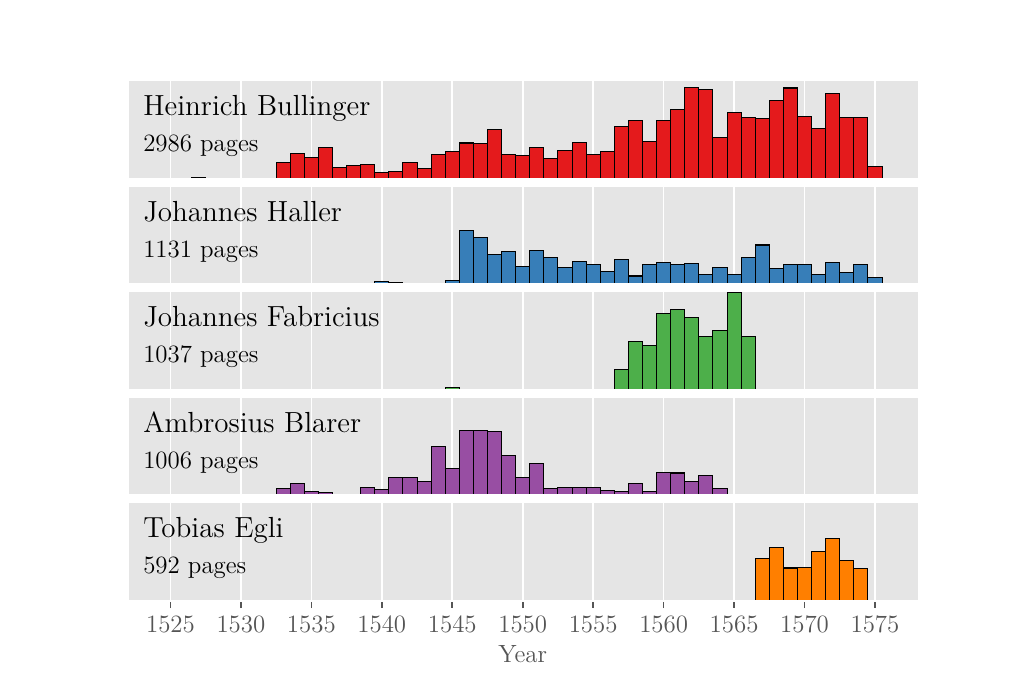}
    \caption{ }
    \label{fig:heatmap}
\end{subfigure}
\hfill
\begin{subfigure}[b]{0.38\linewidth}
    \centering
    \newcommand{\imgwidth}{0.99\linewidth}

    \includegraphics[width=\imgwidth]{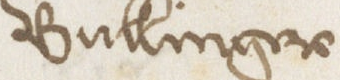}
    \vspace{0.075cm}
    \includegraphics[width=\imgwidth]{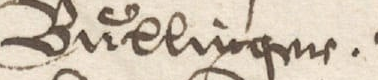}
    \vspace{0.075cm}
    \includegraphics[width=0.45\linewidth]{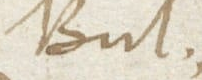}
    \hfill
    \includegraphics[width=0.45\linewidth]{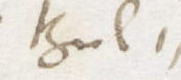}
    \vspace{0.075cm}
    \includegraphics[width=\imgwidth]{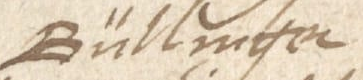}
    \caption{ }
    \label{fig:five_images}
\end{subfigure}

\caption{(a) Activity of the Top-5 writers of the corpus. (b) "Bullinger" written by different individuals.}
\label{fig:combined}
\end{figure}

Spanning 1523–1575, the collection contains about 12,000 preserved letters exchanged with correspondents across the Holy Roman Empire, England, France, Italy, and Eastern Europe. Approximately 2,000 letters were written by Bullinger and around 10,000 were addressed to him \cite{fischer-etal-2022-machine}. Bullinger's writing contributions as well as the four other most frequent ones are shown in Fig. \ref{fig:heatmap}. About 80\% of the texts are in Latin - the primary language of early modern scholarly communication - with most of the remainder in Early New High German and some letters even mixed (\emph{code-switching} \cite{volk-etal-2022-nunc}). Another characteristic is the variety in handwriting style as well as the presence of abbreviations, for example as shown in Fig. \ref{fig:five_images}, where the word "Bullinger" is written as "Bul.".

\subsection{Processing}

In the following, we describe the pipeline to process the digital scans of the letters to make them accessible for the tasks of \ac{WR} and \ac{HTR}. In the course of the project, we processed the full corpus. For the publicly available dataset, we provide a clean version and ensure all samples have validated labels (e.g., empty pages or samples for which the author is unknown are filtered). Finally, we present a split for benchmarking. 

\paragraph{Line Segmentation} The first step in the pipeline is the segmentation of the lines including the layout analysis to enable research for \ac{HTR} systems, usually working on line-level. We use the Transkribus \cite{transkribus} platform to finetune a layout analysis model on 100 manually annotated pages. The layout of the letters is consistent and only includes artifacts of the scans (e.g., archival references as shown in the bottom corners of Fig. \ref{fig:sample}) that should be neglected. Furthermore, we suppress marginalia and footnotes since they are not part of the provided transcriptions for training the transcription model. Hence, the layout - consisting of text regions and  line polygons - is provided in Transkribus' PageXML format. For the cut-out line polygons, we follow \cite{bullinger-dataset2023} and add a randomly generated background based on the texture to obtain rectangular line images.

\paragraph{Text Recognition}

The foundation of our data is the initial Bullinger dataset \cite{bullinger-dataset2023}. For those letters, scholars created transcriptions; however, they were not strictly character-accurate (\emph{semi-diplomatic}). Abbreviations are frequently expanded and punctuation is added, which introduces inconsistencies at the character level. Consequently, the automatically generated ground truth contains a certain degree of noise, posing an additional challenge for training \ac{HTR} systems. Previous analyses have shown that the annotation quality lacks fine-grained detail, particularly with respect to hyphenation due to automatic text-to-image alignment \cite{jungo2023}. To mitigate these issues, we refine the annotations of the first version by following the method proposed in \cite{Peer2025}, resulting in improved accuracy for hyphenations. The final transcriptions are obtained using the best-performing model from \cite{Peer2025}. Evaluation on a subset of 50 manually annotated pages shows that the refined transcriptions achieve a \ac{CER} of 6.5\%. Hence, the ground truth naturally contains errors (or is suboptimal), that cannot be avoided for such a large collection.

\paragraph{Postprocessing} As a last step, we filter pages based on the layout information and exclude samples that contain no handwriting. Secondly, since the corpus also includes 'envelopes' - corresponding to the backside, which usually contains the address of the receiver and may only contain a single word or line (e.g., a signature), we binarize the pages and filter samples that contain black pixels less than 1\% of the text region identified by Transkribus). As a binarization algorithm, we empirically inspect the results of different binarizers for 50 pages and choose Gatos' algorithm \cite{Gatos2004} (similar to Historical-WI \cite{historical-wi}). The binarized images are used for the subsequent task of \ac{WR}.

\paragraph{Writer and Time Information} The meta information used in this paper consists of the writer identity and the date of the writing sample. The initial correspondance comprises different document types: \emph{autographs} (written and sent by the sender), \emph{originals} (sent by the sender but written by another hand), autograph drafts, and autograph copies or partial copies. The identity is confirmed by the signature or - in case of ambiguity due to a name occuring multiple times, e.g., father and son - confirmed by palaeographic analysis, conducted by scholars of the \emph{Heinrich-Bullinger-Briefwechseledition} (HBBW) \cite{HBBW_Reihe}. We restrict the data to autographs for which the writer identity is explicitly documented and is dated for a specific year.

\subsection{Training and Evaluation Splits}
In total, we obtain 20,898 pages by 796 writers. We explain the two splits for future research on both tasks in the following.

\paragraph{Handwritten Text Recognition}

\begin{figure}[htbp]
        \centering
        \input{figs/freq-nonfreq}
        \caption{Distribution of the number of letters per writer.}
        \label{fig:freq-nonfreq}
\end{figure}
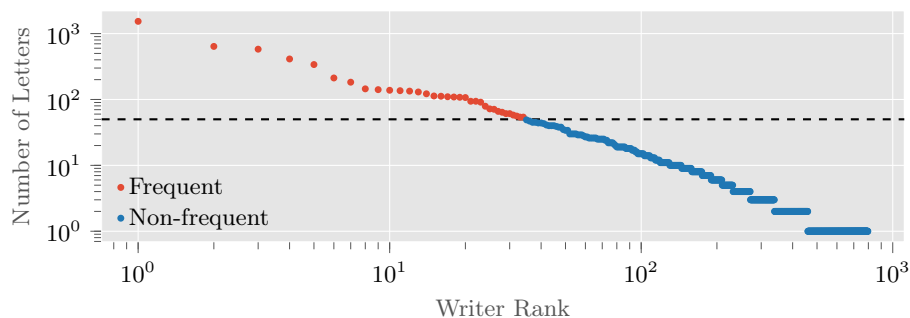

The database is split following the protocol proposed by Scius-Bertrand et al. \cite{bullinger-dataset2023}. We first sort all writers by their number of letters, which results in a Zipf-like distribution (see Fig.~\ref{fig:freq-nonfreq}, for better visibility we show a log-log plot). Using a threshold of 50 letters, we distinguish frequent writers with at least 50 letters and non-frequent writers with fewer than 50 letters.
For frequent writers, we split their letters writer-wise into 90\% training (for which we reserve 10\% for validation), and 10\% testing. This yields 144 writers with 376,582 lines in training and a corresponding frequent-writer test set with 939 letters (48,197 lines). All material from non-frequent writers is reserved for a separate test set, comprising 652 writers with 74,443 lines. This setup enables us to evaluate performance on known writers (frequent test set with unseen letters, but writing style is seen during training) and unknown writers (non-frequent test set, unseen handwriting style during training), reflecting a scenario used by scholars. The statistics are shown in Table \ref{tab:htr-stats}.

\paragraph{Writer Retrieval} We provide a writer-disjoint train/test-split with a ratio of  20:80 that is orientated at CVL \cite{cvl} and Historical-WI \cite{historical-wi}. This enables the \emph{leave-one-out-validation} scheme as commonly applied by \ac{WR} methods for which each sample of the test set is once used as a query in an open-set scenario. Furthermore, we automatically split the dataset to minimize  the following aspects to ensure compatibility with related work and also enable research on currently unexplored tasks, while each writer is only present in either test or train:
\begin{itemize}
    \item The difference in the time distribution between training and test set should be minimized.
    \item The intra-writer variance with respect to the year should be balanced (both splits should include writers who contributed over a longer period of time).
\end{itemize}

We also assign Heinrich Bullinger - the writer with the most documents in the corpus - to the test split. Given this procedure, we obtain a train/test split of 4,396/16,502 pages written by 164/631 individuals, as shown in Table \ref{tab:wr-stats} (time distribution plots are provided in the preview).

\begin{minipage}[t]{0.55\textwidth}
    \centering
    \captionof{table}{Dataset statistics for \ac{HTR}}
    \label{tab:htr-stats}
    \begin{tabular}{lccc}
    \toprule
     & Training & \makecell{Test \\ Frequent} & \makecell{Test \\ Non-Frequent}  \\
    \midrule
    Writers & 144 & 144 & 652  \\
    Letters & 7,834 & 939 & 1,529  \\
    Pages   & 15,619 & 1,957 & 3,311 \\
    Lines   & 376,582 & 48,197 & 74,443  \\
    \bottomrule
    \end{tabular}
\end{minipage}
\hfill
\begin{minipage}[t]{0.42\textwidth}
    \centering
    \captionof{table}{Dataset statistics for \ac{WR}}
    \label{tab:wr-stats}
    \begin{tabular}{lcc}
    \toprule
     & Training & Test  \\
    \midrule
    Writers & 164 & 632  \\
    Letters & 2,142 & 8,160 \\
    Pages   & 4,396 & 16,502 \\
    \bottomrule
    \end{tabular}
\end{minipage}






%% file: figs/freq-nonfreq.tex
\begin{tikzpicture}

\definecolor{chocolate2267451}{RGB}{226,74,51}
\definecolor{nonfreqblue}{RGB}{31,119,180}
\definecolor{dimgray85}{RGB}{85,85,85}
\definecolor{gainsboro229}{RGB}{229,229,229}

\begin{axis}[
axis background/.style={fill=gainsboro229},
axis line style={white},
log basis x={10},
log basis y={10},
tick align=outside,
tick pos=left,
x grid style={white},
y grid style={white},
xmajorgrids,
ymajorgrids,
width=\textwidth,
height=0.38\textwidth,
xlabel=\textcolor{dimgray85}{Writer Rank},
ylabel=\textcolor{dimgray85}{Number of Letters},
xmin=0.716068132595609, xmax=1111.62606428896,
ymin=0.692938323521023, ymax=2215.20436653036,
xmode=log,
ymode=log,
xtick style={color=dimgray85},
ytick style={color=dimgray85},
legend style={
    at={(0.01,0.01)},
    anchor=south west,
    draw=none,
    fill=none,
    font=\footnotesize
},
legend cell align=left
]

\addplot [
    only marks,
    mark=*,
    mark size=1,
    semithick,
    chocolate2267451
]
table {
1 1535
2 639
3 580
4 412
5 339
6 212
7 183
8 145
9 141
10 138
11 136
12 134
13 130
14 122
15 113
16 112
17 110
18 109
19 108
20 107
21 94
22 94
23 91
24 79
25 72
26 71
27 66
28 64
29 61
30 61
31 58
32 56
33 54
34 54
};
\addlegendentry{Frequent}

\addplot [
    only marks,
    mark=*,
    mark size=1,
    semithick,
    nonfreqblue
]
table {
35 49
36 47
37 45
38 45
39 44
40 44
41 43
42 41
43 40
44 40
45 40
46 39
47 38
48 38
49 35
50 34
51 34
52 30
53 30
54 30
55 30
56 29
57 29
58 29
59 28
60 27
61 27
62 26
63 26
64 26
65 26
66 26
67 25
68 25
69 25
70 25
71 25
72 24
73 24
74 22
75 22
76 22
77 22
78 21
79 20
80 19
81 19
82 19
83 19
84 19
85 19
86 19
87 18
88 18
89 18
90 18
91 18
92 17
93 17
94 17
95 16
96 16
97 15
98 15
99 15
100 15
101 15
102 15
103 14
104 14
105 14
106 14
107 14
108 14
109 13
110 13
111 13
112 13
113 13
114 12
115 12
116 12
117 12
118 12
119 11
120 11
121 11
122 11
123 11
124 11
125 11
126 11
127 11
128 11
129 11
130 10
131 10
132 10
133 10
134 10
135 10
136 10
137 10
138 10
139 10
140 10
141 10
142 10
143 10
144 10
145 9
146 9
147 9
148 9
149 9
150 9
151 9
152 9
153 9
154 9
155 9
156 9
157 9
158 9
159 8
160 8
161 8
162 8
163 8
164 8
165 8
166 8
167 8
168 8
169 8
170 8
171 8
172 8
173 8
174 8
175 7
176 7
177 7
178 7
179 7
180 7
181 7
182 7
183 7
184 7
185 7
186 7
187 7
188 7
189 7
190 6
191 6
192 6
193 6
194 6
195 6
196 6
197 6
198 6
199 6
200 6
201 6
202 6
203 6
204 6
205 6
206 6
207 6
208 6
209 6
210 5
211 5
212 5
213 5
214 5
215 5
216 5
217 5
218 5
219 5
220 5
221 5
222 5
223 5
224 5
225 5
226 5
227 5
228 5
229 5
230 5
231 5
232 4
233 4
234 4
235 4
236 4
237 4
238 4
239 4
240 4
241 4
242 4
243 4
244 4
245 4
246 4
247 4
248 4
249 4
250 4
251 4
252 4
253 4
254 4
255 4
256 4
257 4
258 4
259 4
260 4
261 4
262 4
263 4
264 4
265 4
266 4
267 4
268 4
269 4
270 4
271 4
272 3
273 3
274 3
275 3
276 3
277 3
278 3
279 3
280 3
281 3
282 3
283 3
284 3
285 3
286 3
287 3
288 3
289 3
290 3
291 3
292 3
293 3
294 3
295 3
296 3
297 3
298 3
299 3
300 3
301 3
302 3
303 3
304 3
305 3
306 3
307 3
308 3
309 3
310 3
311 3
312 3
313 3
314 3
315 3
316 3
317 3
318 3
319 3
320 3
321 3
322 3
323 3
324 3
325 3
326 3
327 3
328 3
329 3
330 3
331 3
332 3
333 3
334 3
335 3
336 3
337 3
338 3
339 2
340 2
341 2
342 2
343 2
344 2
345 2
346 2
347 2
348 2
349 2
350 2
351 2
352 2
353 2
354 2
355 2
356 2
357 2
358 2
359 2
360 2
361 2
362 2
363 2
364 2
365 2
366 2
367 2
368 2
369 2
370 2
371 2
372 2
373 2
374 2
375 2
376 2
377 2
378 2
379 2
380 2
381 2
382 2
383 2
384 2
385 2
386 2
387 2
388 2
389 2
390 2
391 2
392 2
393 2
394 2
395 2
396 2
397 2
398 2
399 2
400 2
401 2
402 2
403 2
404 2
405 2
406 2
407 2
408 2
409 2
410 2
411 2
412 2
413 2
414 2
415 2
416 2
417 2
418 2
419 2
420 2
421 2
422 2
423 2
424 2
425 2
426 2
427 2
428 2
429 2
430 2
431 2
432 2
433 2
434 2
435 2
436 2
437 2
438 2
439 2
440 2
441 2
442 2
443 2
444 2
445 2
446 2
447 2
448 2
449 2
450 2
451 2
452 2
453 2
454 2
455 2
456 2
457 2
458 2
459 2
460 1
461 1
462 1
463 1
464 1
465 1
466 1
467 1
468 1
469 1
470 1
471 1
472 1
473 1
474 1
475 1
476 1
477 1
478 1
479 1
480 1
481 1
482 1
483 1
484 1
485 1
486 1
487 1
488 1
489 1
490 1
491 1
492 1
493 1
494 1
495 1
496 1
497 1
498 1
499 1
500 1
501 1
502 1
503 1
504 1
505 1
506 1
507 1
508 1
509 1
510 1
511 1
512 1
513 1
514 1
515 1
516 1
517 1
518 1
519 1
520 1
521 1
522 1
523 1
524 1
525 1
526 1
527 1
528 1
529 1
530 1
531 1
532 1
533 1
534 1
535 1
536 1
537 1
538 1
539 1
540 1
541 1
542 1
543 1
544 1
545 1
546 1
547 1
548 1
549 1
550 1
551 1
552 1
553 1
554 1
555 1
556 1
557 1
558 1
559 1
560 1
561 1
562 1
563 1
564 1
565 1
566 1
567 1
568 1
569 1
570 1
571 1
572 1
573 1
574 1
575 1
576 1
577 1
578 1
579 1
580 1
581 1
582 1
583 1
584 1
585 1
586 1
587 1
588 1
589 1
590 1
591 1
592 1
593 1
594 1
595 1
596 1
597 1
598 1
599 1
600 1
601 1
602 1
603 1
604 1
605 1
606 1
607 1
608 1
609 1
610 1
611 1
612 1
613 1
614 1
615 1
616 1
617 1
618 1
619 1
620 1
621 1
622 1
623 1
624 1
625 1
626 1
627 1
628 1
629 1
630 1
631 1
632 1
633 1
634 1
635 1
636 1
637 1
638 1
639 1
640 1
641 1
642 1
643 1
644 1
645 1
646 1
647 1
648 1
649 1
650 1
651 1
652 1
653 1
654 1
655 1
656 1
657 1
658 1
659 1
660 1
661 1
662 1
663 1
664 1
665 1
666 1
667 1
668 1
669 1
670 1
671 1
672 1
673 1
674 1
675 1
676 1
677 1
678 1
679 1
680 1
681 1
682 1
683 1
684 1
685 1
686 1
687 1
688 1
689 1
690 1
691 1
692 1
693 1
694 1
695 1
696 1
697 1
698 1
699 1
700 1
701 1
702 1
703 1
704 1
705 1
706 1
707 1
708 1
709 1
710 1
711 1
712 1
713 1
714 1
715 1
716 1
717 1
718 1
719 1
720 1
721 1
722 1
723 1
724 1
725 1
726 1
727 1
728 1
729 1
730 1
731 1
732 1
733 1
734 1
735 1
736 1
737 1
738 1
739 1
740 1
741 1
742 1
743 1
744 1
745 1
746 1
747 1
748 1
749 1
750 1
751 1
752 1
753 1
754 1
755 1
756 1
757 1
758 1
759 1
760 1
761 1
762 1
763 1
764 1
765 1
766 1
767 1
768 1
769 1
770 1
771 1
772 1
773 1
774 1
775 1
776 1
777 1
778 1
779 1
780 1
781 1
782 1
783 1
784 1
785 1
786 1
787 1
788 1
789 1
790 1
791 1
792 1
793 1
794 1
795 1
796 1
};
\addlegendentry{Non-frequent}
\addplot [
    thick,
    dashed,
    black
] coordinates {
    (0.716068132595609, 50)
    (1111.62606428896, 50)
};
\end{axis}
\end{tikzpicture}

%% file: sections/4_methods.tex
\section{Methods}\label{sec:methods}

In this section, we briefly describe the methods and the metrics used for baseline evaluation in the domain of \ac{HTR} and \ac{WR}.

\subsection{Handwritten Text Recognition}

We evaluate four systems that are aligned with state of the art and also represent different architectures. We use the default parameters for the systems except mentioned otherwise:

 HTRFlor~\cite{deSousaNeto2020} adopts a compact CNN-based architecture followed by stacked bidirectional GRU layers for sequence modeling. The model emphasizes architectural simplicity and efficiency while maintaining competitive recognition performance. Compared to PyLaia and Retsinas et al., it replaces BiLSTMs with BGRUs, that results in a lighter recurrent design.
 
 PyLaia~\cite{PyLaia} follows a conventional CNN--BiLSTM--CTC pipeline for handwritten text recognition. A deep convolutional backbone extracts visual features that are processed by stacked BiLSTM layers and optimized using CTC loss. Unlike Retsinas et al., it does not include an auxiliary shortcut branch and relies on standard pooling operations.

Retsinas et al.~\cite{Retsinas2022} propose a CNN-based architecture that preserves the input image aspect ratio and replaces max-pooling with column-wise concatenation to reduce information loss. A distinctive feature is the introduction of a parallel shortcut branch with a 1D convolution trained with CTC loss, which guides the CNN backbone toward learning more discriminative features. Sequence modeling is performed using stacked BiLSTM layers. We remove brightness and contrast data augmentation for our training since we have samples with low contrast.

 TrOCR~\cite{trocr} departs from the CNN--RNN--CTC paradigm by employing a Transformer-based encoder--decoder architecture. It performs autoregressive sequence generation and does not rely on recurrent layers or CTC loss. This design aims at end-to-end training with large-scale pretraining. We finetune the base variant for 15 epochs.

\subsubsection*{Metrics} We follow related work and evaluate \ac{HTR} systems via the edit-distance-based \ac{CER} and \ac{WER} metrics.

\subsection{Writer Retrieval}

To benchmark \bull \ for \ac{WR}, we evaluate state-of-the-art pipelines spanning CNN-based aggregation models, transformer-based global descriptors, and attention-driven masking approaches. For all methods, we follow the schemes proposed in the original work.

In line with prior work, binarized document images are decomposed into $32\times32$ patches detected at SIFT keypoints to capture local handwriting characteristics while ensuring sufficient training samples. We consider both a supervised setting (\emph{Sup.}), where writer labels are used during training, and an unsupervised clustering-based surrogate supervision setting (\emph{Cl-S.}) \cite{unsupervised_icdar17}, where k-means clustering (with 5k centers) on the corresponding SIFT descriptors provides pseudo-labels for representation learning.

First, we reproduce the CNN-based aggregation framework of Christlein et al. \cite{unsupervised_icdar17} by using ResNet56 (following \cite{peer_netrvlad}). Furthermore, we also train the vanilla vision transformer as used in \cite{peer2024} that has shown to be superior for complex datasets. We evaluate different aggregation schemes of the local patch embeddings: Sum-Pooling, mVLAD with five vocabularies, and the learned variant NetVLAD as proposed in \cite{peer_netrvlad} for \ac{WR}. We use 2000 patches per sample for aggregation.

Finally, we assess the self-supervised AttMask model proposed by Raven et al. \cite{raven}, which uses a ViT backbone with attention-guided masking to suppress non-informative regions and directly produces global writer descriptors, trained on $224\times 224$ crops. We only aggregate foreground tokens as proposed in the original work.

For all codebooks, we use a vocabulary size of 100 clusters. The page descriptors are powernormalized and whitened. If the dimension is higher, it is reduced to 512 via PCA. The PCA is fitted on the corresponding training descriptors.

\subsubsection*{Metrics}

\ac{WR} is evaluated in a \emph{leave-one-out} scenario: For each query document $q$, all remaining $N-1$ documents are ranked by cosine distance in the embedding space. Documents authored by the same writer as $q$ are considered relevant. Apart from the commonly used Top-$n$ accuracies (we use the soft criterion) and \ac{mAP}, we introduce the temporal \ac{nDCG} to evaluate \ac{WR} approaches based on chronological coherence. The metric is usually applied in information retrieval (e.g., for recommender systems). Instead of a binary relevance indicator as used for calculating \ac{mAP}, we define graded relevance using document year $y_k$ and query year $y_q$ with their writers $w_k$ and $w_q$:

\begin{equation*}
        rel_q(k) = \left\{\begin{array}{lr}
        1 - \frac{\lvert y_k - y_q\rvert}{T_\mathrm{max}}, & \ \ \ \ \mathrm{if} \ w_k = w_q\\
        0 & \ \ \ \ \mathrm{otherwise} \\
        \end{array} \right.
\end{equation*}

where $T_\mathrm{max}$ is the time window (for the test split of \bull \ $T_\mathrm{max} = 51$ years, corresponding to 1524--1575). The DCG is then defined as

\[
DCG_q= \sum_{k=1}^{N-1} \frac{2^{rel_q(k)} - 1}{\log_2(k+1)} .
\]

Finally, the \ac{nDCG} for a query $q$ is computed by normalizing with the ideal ranking $IDCG_q$

\[
nDCG_q = \frac{DCG_q}{IDCG_q}.
\]

where $IDCG_q$ is obtained by sorting the gallery by decreasing relevance. The final performance is reported as the mean nDCG over all queries.

%% file: sections/5_experiments.tex
\section{Benchmarks}\label{sec:results}
 We provide an evaluation of the state-of-the-art methods including an analysis on \bull \ in the following.
 

\subsection{Handwritten Text Recognition}

Our four \ac{HTR} systems differ not only in their architectures, but also in their number of parameters. We train all models on the training set in their default settings and report the results on the subsets of (non-)frequent writers as well as the overall performance. As shown in Table \ref{tab:htr-results}, the models range from less than 1M to 334M parameters. The results show a clear improvement when moving from the smallest model to the medium-sized architectures, while the gap between the medium-scale models and the largest transformer-based model is comparatively smaller in relation to the large increase in parameter count. Although the largest model achieves the best overall performance (particularly for WER), PyLaia attains similar CERs despite being considerably smaller. Across all systems, performance on non-frequent writers is consistently worse than on frequent writers, and this difference persists independently of model size.

\begin{table}[h]
\centering
\caption{Error rates for \bull \ on the two subsets proposed. Best performance is highlighted in \textbf{bold}, second best \underline{underlined}.}\label{tab:htr-results}

\begin{tabular}{lccccccc}
\toprule
  &  Params (M) & \multicolumn{2}{c}{Frequent} 
  & \multicolumn{2}{c}{Non-Frequent} & \multicolumn{2}{c}{Total} \\ 
  & & CER  & WER & CER & WER & CER  & WER \\
\midrule
HTRFlor \cite{deSousaNeto2020} & 0.8 & 17.9 & 50.8 & 19.4 & 54.5 & 19.2 & 54.0\\
PyLaia \cite{PyLaia} & 4.8 & \underline{8.3} & \underline{30.2} & \underline{9.7} & {34.9} & \underline{9.2} & {33.1} \\
Retsinas et al. \cite{Retsinas2022} & 7.4 & 9.2 & 30.3 & {10.2} & \underline{34.1} & {9.8} & \underline{32.4}  \\
TrOCR \cite{trocr} & 334 & \textbf{8.2} & \textbf{27.9} & \textbf{9.6} & \textbf{32.2} & \textbf{9.1} & \textbf{30.6} \\
\bottomrule
\end{tabular}
\end{table}

\begin{figure}[h]
\centering

\begin{subfigure}{\textwidth}
\centering
\input{figs/cer/freq/plot_1_cer_vs_length}
\hfill
\input{figs/cer/freq/plot_2_mean_cer_per_writer}

\caption{Frequent writers}
\end{subfigure}

\vspace{0.8em}

\begin{subfigure}{\textwidth}
\centering
\input{figs/cer/nonfreq/plot_1_cer_vs_length}
\hfill
\input{figs/cer/nonfreq/plot_2_mean_cer_per_writer}

\caption{Non-frequent writers}
\end{subfigure}

\vspace{0.1em}

\caption{Error analysis for (a) frequent and (b) non-frequent writers. Writers are sorted according to their frequency, IDs for the writers with the highest three CER values are shown.}
\label{fig:cer_analysis}
\end{figure}
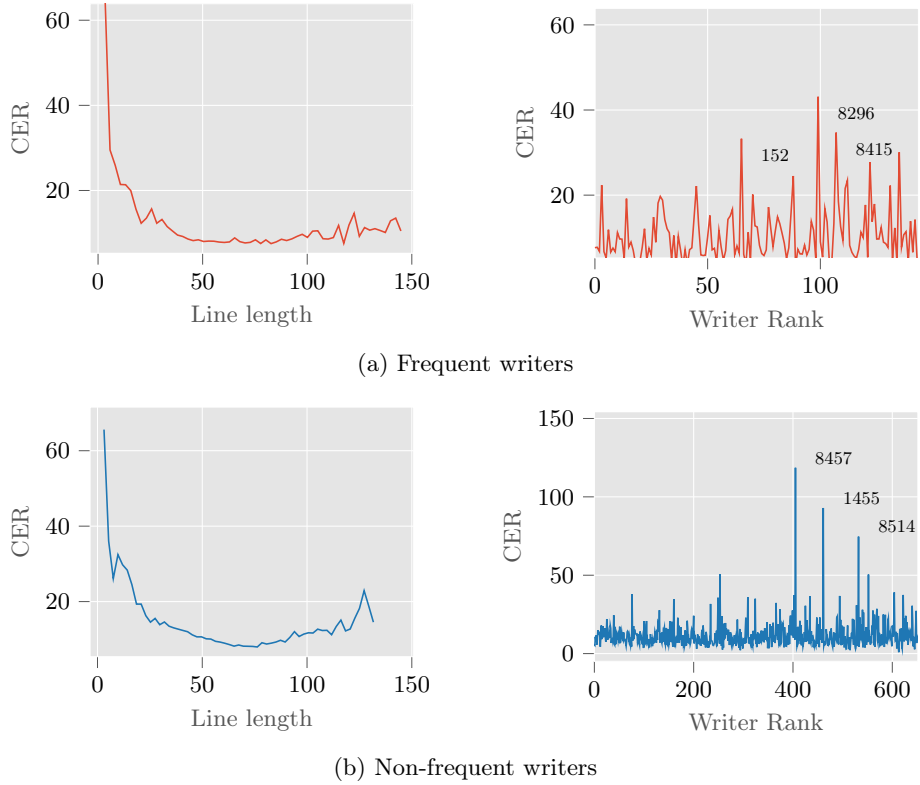

\begin{figure}
\centering
\begin{subfigure}{0.95\textwidth}
    \includegraphics[width=0.8\linewidth]{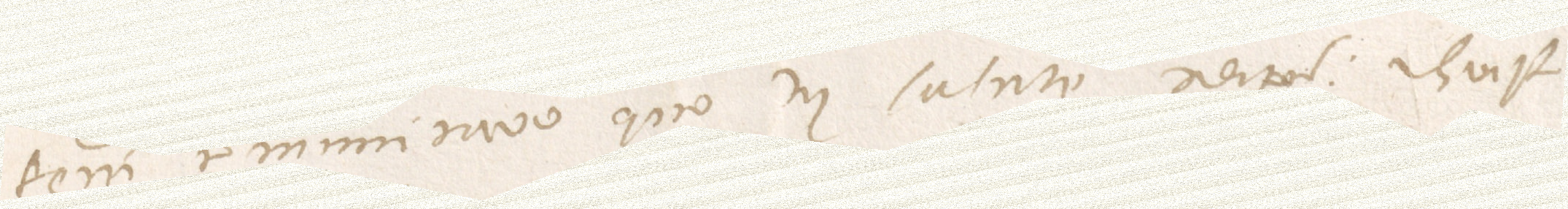}
    \caption*{\small \textbf{Pred}: \texttt{ponimminuro, quo in salute adit.} (ID 8296, CER: 52.1\%)\\ \textbf{GT}: \texttt{nicommunitare, quo in salute reipublicae Christi} \\
    }
    \end{subfigure}

\begin{subfigure}{0.95\textwidth}
    \includegraphics[width=0.8\linewidth]{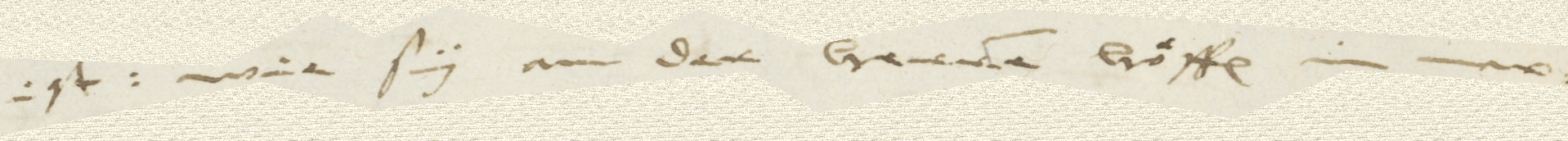}
    \caption*{\small \textbf{Pred}: \texttt{ist, wie sy an der herren höffen in mar} (ID 495, CER: 0.0\%)\\ \textbf{GT}: \texttt{ist, wie sy an der herren höffen in mar} \\
    }

\end{subfigure}
\caption{Qualitative Examples}\label{fig:qual-htr}

\end{figure}

\paragraph{Error analysis} Next, we analyze the \ac{CER} distributions of the best-performing TrOCR model to provide insights in the failure cases. Fig. \ref{fig:cer_analysis} presents the \ac{CER} analysis for (a) frequent and (b) non-frequent writers. The left column shows the relationship between CER and line length, while the right column reports the mean CER per writer, with writers ranked by frequency. 
For frequent writers, CER decreases sharply as line length increases and stabilizes for medium-length lines, suggesting that very short lines are more error-prone. In \bull, shorter lines usually refer to headers or adresses on the backside, which might be more difficult for the model due to non-language parts. The per-writer distribution is relatively compact, with only a few noticeable peaks corresponding to the three highest-CER writers. In contrast, non-frequent writers (bottom row) exhibit greater variability: although CER follows a similar decreasing trend with increasing line length, the writer ranking shows three writers that have a \ac{CER} of at least 50\%. Two qualitative examples for a writer with a high CER and Bullinger (495) are shown in Fig. \ref{fig:qual-htr}.

\subsection{Writer Retrieval}

Our results for \ac{WR} are presented in Table~\ref{tab:writer_retrieval}. 
AttMask \cite{raven} (trained on $224\times 224$ patches) outperforms all $32\times 32$ patch-based models with both encodings (SumPooling and VLAD). Due to the large dataset size, the SSL training protocol seems to perform as a better feature extractor with the disadvantage of being computationally expensive, with training times of about three days, compared to the other models (2--3h on a RTX 6000).

\begin{table}[t]
\centering
\caption{Benchmark for \ac{WR} on \bull. Best performance is highlighted in \textbf{bold}, second best \underline{underlined}. ViT-architectures are denoted with the patch size and depth.}
\label{tab:writer_retrieval}
\begin{tabular}{lllccccc}
\toprule
Model & Training & Encoding & mAP & nDCG & Top-1 & Top-5& Top-10  \\
\midrule
\multirow{8}{*}{ResNet56}
 & \multirow{4}{*}{Sup.}
   & Sum-Pooling   & 49.9 & 78.2 & 90.1 & 94.0 & 95.3   \\
   & & VLAD & 48.7 & 76.6 & 83.1 & 91.6 & 93.6 \\
   & & mVLAD & 46.2 & 76.3 & 88.6 & 93.7 & 95.0 \\
   & & NetRVLAD & 51.7 & 79.3 & 91.0 & 94.9 & 96.0 \\ \cmidrule(l){2-8}
 & \multirow{4}{*}{Cl-S.}
   & Sum-Pooling  &  51.9 & 78.3 & 86.5 & 92.7 & 94.4  \\ 
  & & VLAD & 48.7 & 76.6 & 83.1 & 91.6 & 93.6 \\
   & & mVLAD & 57.7 & 81.2 & 89.0 & 94.6 & 95.8 \\
   & & NetRVLAD & {65.1} & {84.8} & \underline{93.4} & {96.3} & {97.0}
 \\ \midrule
\multirow{8}{*}{ViT-4/8}
 & \multirow{3}{*}{Sup.}
   & Sum-Pooling & 49.7 & 75.6 & 82.6 & 90.4 & 90.5\\ 
   & & mVLAD & 50.2 & 76.9 & 86.1 & 91.9 & 92.9
 \\
    & & VLAD & 49.6 & 75.2 & 82.4 & 90.1 & 90.4 \\
   & & NetRVLAD & 50.7 & 78.1 & 89.5 & 93.4 & 94.8
 \\ \cmidrule(l){2-8}
   & \multirow{4}{*}{Cl-S.}
   & Sum-Pooling   & 51.8 & 78.4 & 87.0 & 93.2 & 94.8  \\ 
   & & VLAD & 49.6 & 77.0 & 83.6 & 91.9 & 94.0 \\
   & & mVLAD & 55.5 & 80.1 & 87.7 & 93.9 & 95.2 \\
   & & NetRVLAD & {61.9} & {83.4} & {92.5} & \underline{95.6} & \underline{96.4}
\\ \midrule
\multirow{2}{*}{ViT-16/12} & \multirow{2}{*}{AttMask} & SumPooling & \underline{68.1} &  \underline{85.7}& {92.8} & \underline{95.6} & {96.1}\\
&  & VLAD & \textbf{78.3} & \textbf{90.3} & \textbf{95.6} & \textbf{97.5} & \textbf{98.0}\\
\bottomrule
\end{tabular}
\end{table}

Across the encoding configurations, NetRVLAD \cite{peer_netrvlad} consistently yields the strongest retrieval performance, followed by mVLAD \cite{unsupervised_icdar17}, while simple sum-pooling and VLAD perform weaker. We also notice that the unsupervised setting has better scores, which might be due to the temporal aspect of our data that makes supervised training noisier.

Furthermore, ViT-4/8 architecture remains competitive but does not surpass the strongest ResNet-based configuration. Although Top-5 and Top-10 accuracies are generally high (indicating that the correct writer is usually retrieved within the first few candidates), the mAP values remain lower than related work (Historical-WI: ~80\% , HisIR >90\% \cite{raven}). This suggests that - while early ranking positions are reliable - the overall ranking quality across the full retrieval list still leaves room for improvement. At the same time, the relatively high nDCG scores show that temporal coherency within the ranking is well preserved, as we illustrate in qualitative retrieval results in Fig. \ref{fig:wr-viz}.

\definecolor{chocolate2267451}{RGB}{226,74,51}
\definecolor{nonfreqblue}{RGB}{31,119,180}
\definecolor{gggreen}{HTML}{00BA38}
\begin{figure}[t]
\centering

\begin{subfigure}[t]{0.32\textwidth}
\centering

\fcolorbox{nonfreqblue}{white}{
\includegraphics[width=0.9\linewidth]{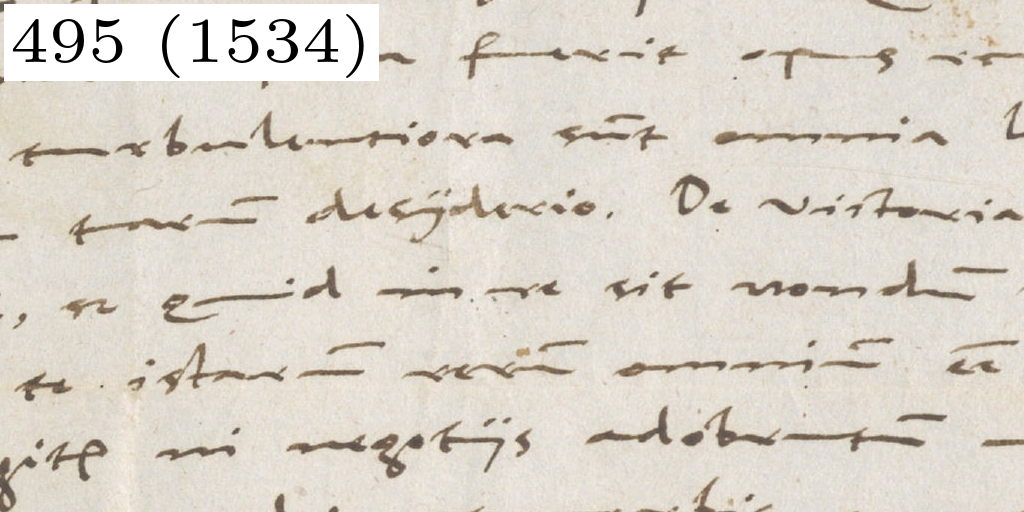}
}

\vspace{0.25mm}
\rule{\linewidth}{0.4pt}
\vspace{0.25mm}

\fcolorbox{gggreen}{white}{
\includegraphics[width=0.9\linewidth]{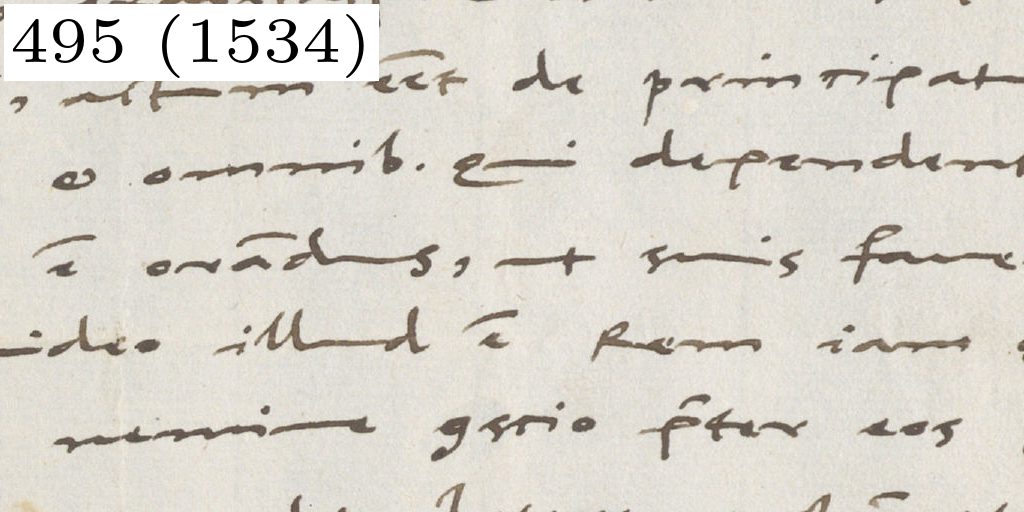}
}

\vspace{1mm}

\fcolorbox{gggreen}{white}{
\includegraphics[width=0.9\linewidth]{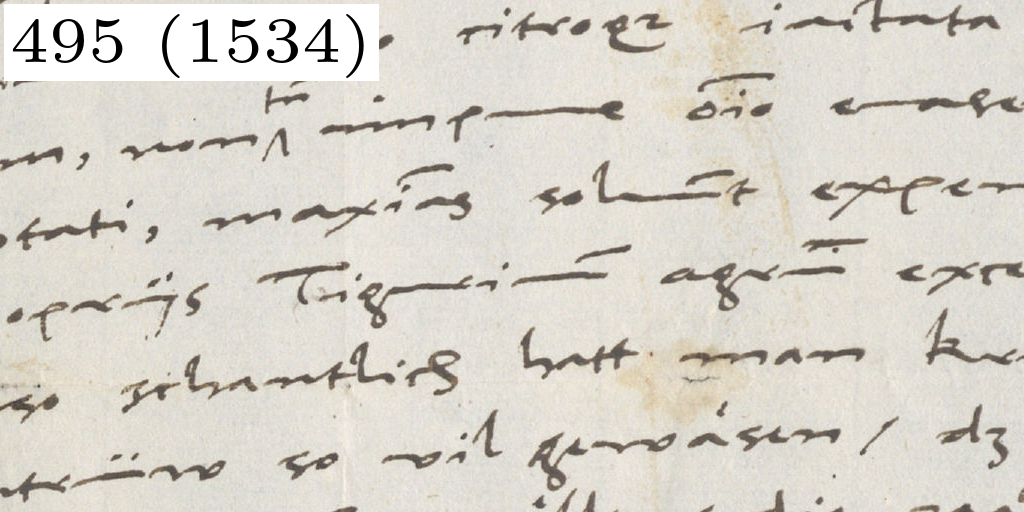}
}

\vspace{1mm}

\fcolorbox{gggreen}{white}{
\includegraphics[width=0.9\linewidth]{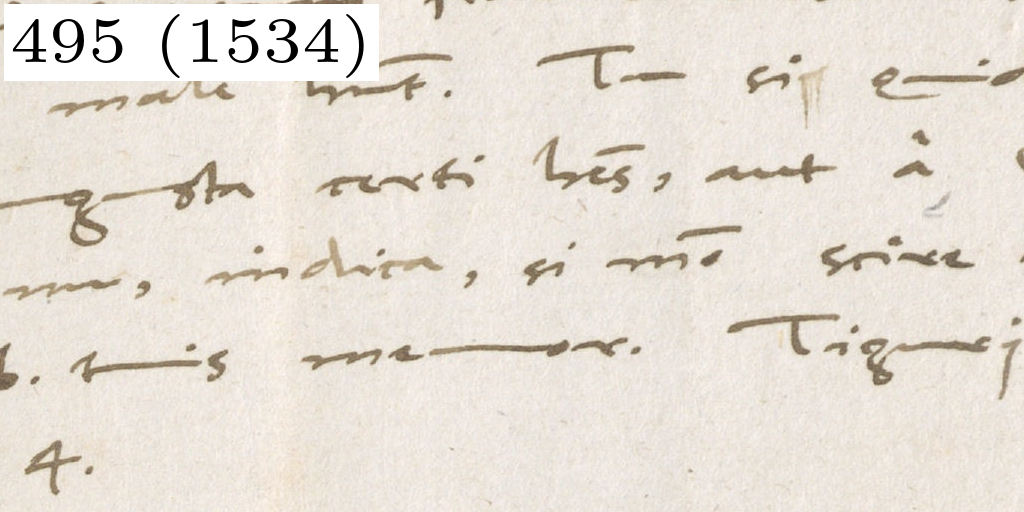}
}

\vspace{1mm}

\fcolorbox{gggreen}{white}{
\includegraphics[width=0.9\linewidth]{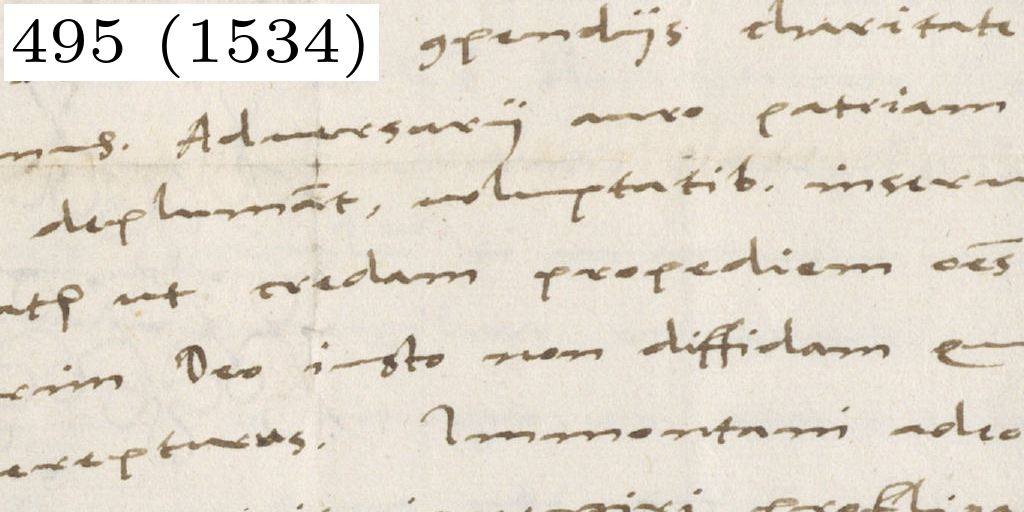}
}
\caption{ }

\end{subfigure}
\hfill
\begin{subfigure}[t]{0.32\textwidth}
\centering

\fcolorbox{nonfreqblue}{white}{
\includegraphics[width=0.9\linewidth]{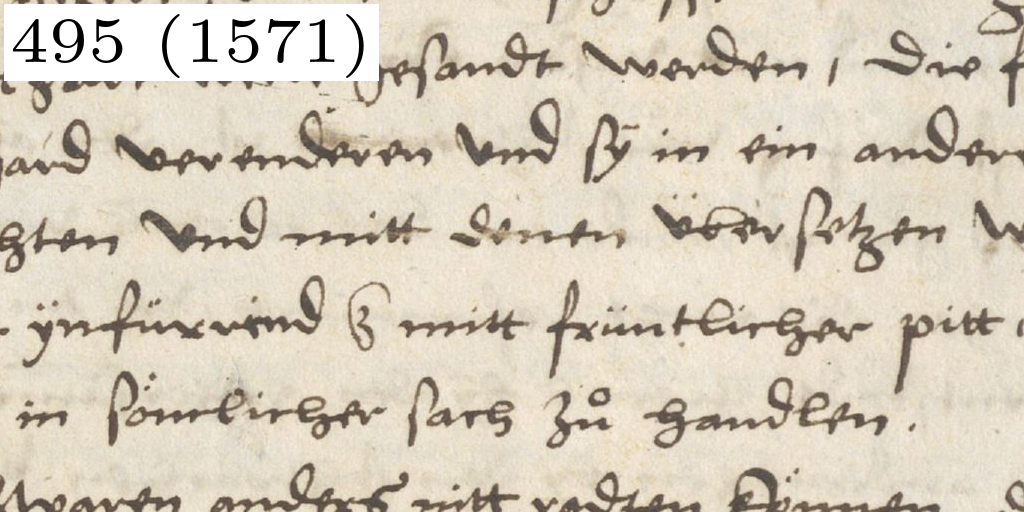}
}

\vspace{0.5mm}
\rule{\linewidth}{0.4pt}
\vspace{0.5mm}

\fcolorbox{gggreen}{white}{
\includegraphics[width=0.9\linewidth]{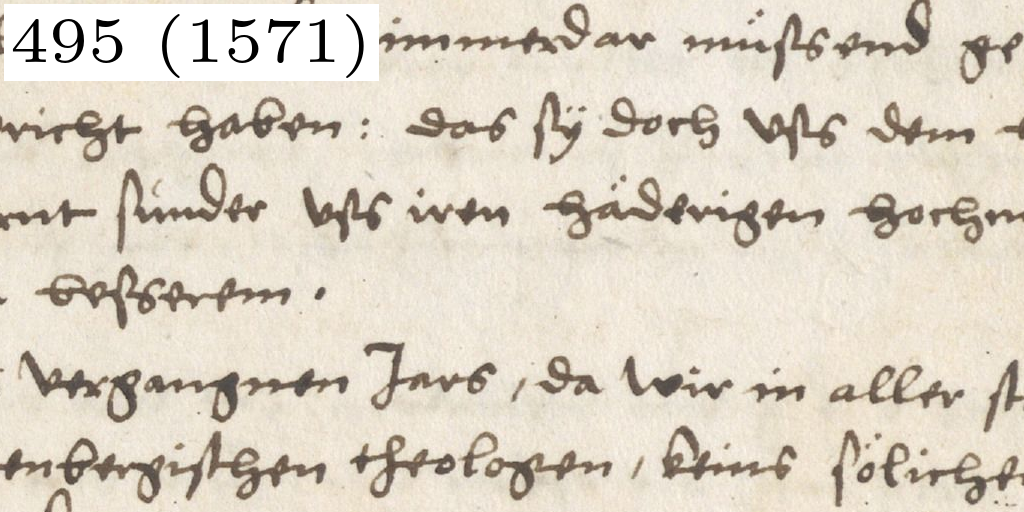}
}

\vspace{1mm}

\fcolorbox{gggreen}{white}{
\includegraphics[width=0.9\linewidth]{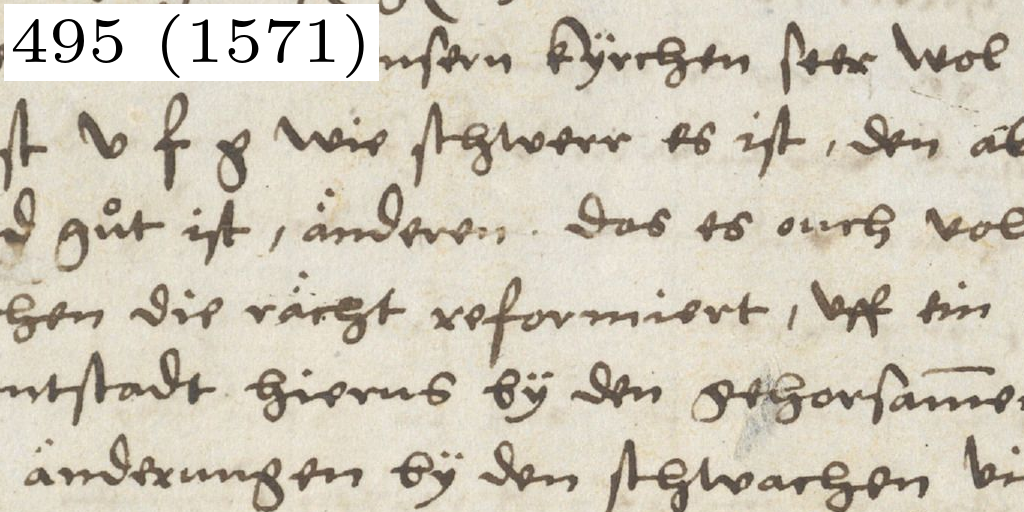}
}

\vspace{1mm}

\fcolorbox{gggreen}{white}{
\includegraphics[width=0.9\linewidth]{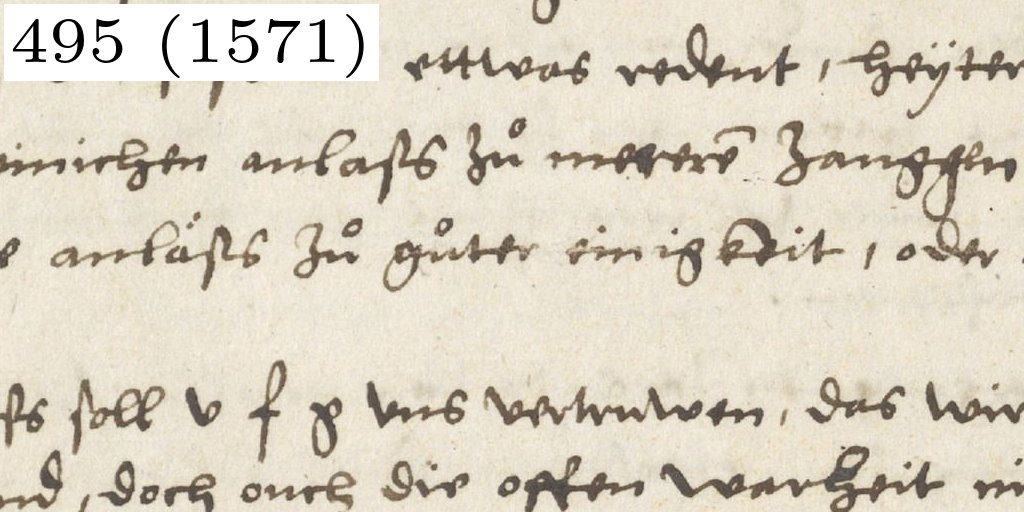}
}

\vspace{1mm}

\fcolorbox{chocolate2267451}{white}{
\includegraphics[width=0.9\linewidth]{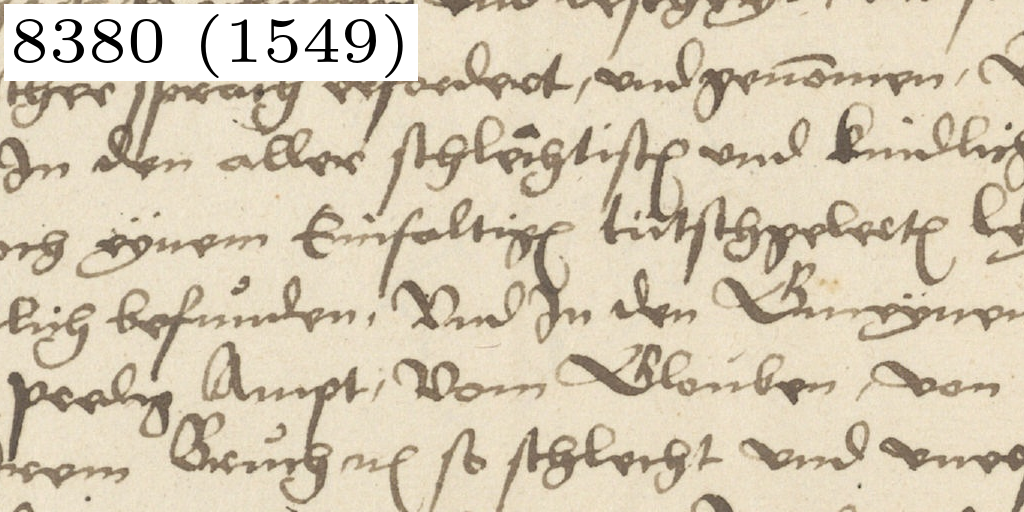}
}
\caption{ }

\end{subfigure}
\hfill
\begin{subfigure}[t]{0.32\textwidth}
\centering

\fcolorbox{nonfreqblue}{white}{
\includegraphics[width=0.9\linewidth]{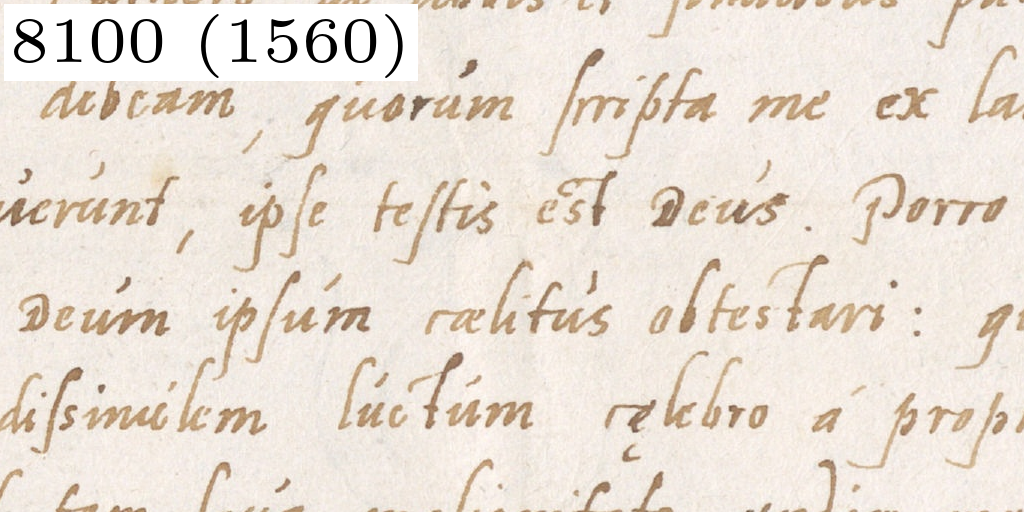}
}

\vspace{0.5mm}
\rule{\linewidth}{0.4pt}
\vspace{0.5mm}

\fcolorbox{gggreen}{white}{
\includegraphics[width=0.9\linewidth]{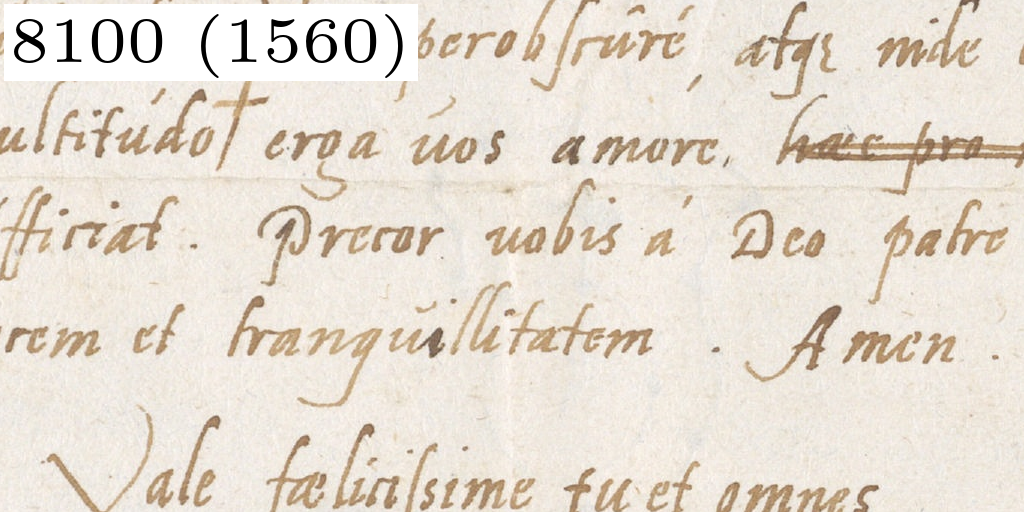}
}

\vspace{1mm}

\fcolorbox{chocolate2267451}{white}{
\includegraphics[width=0.9\linewidth]{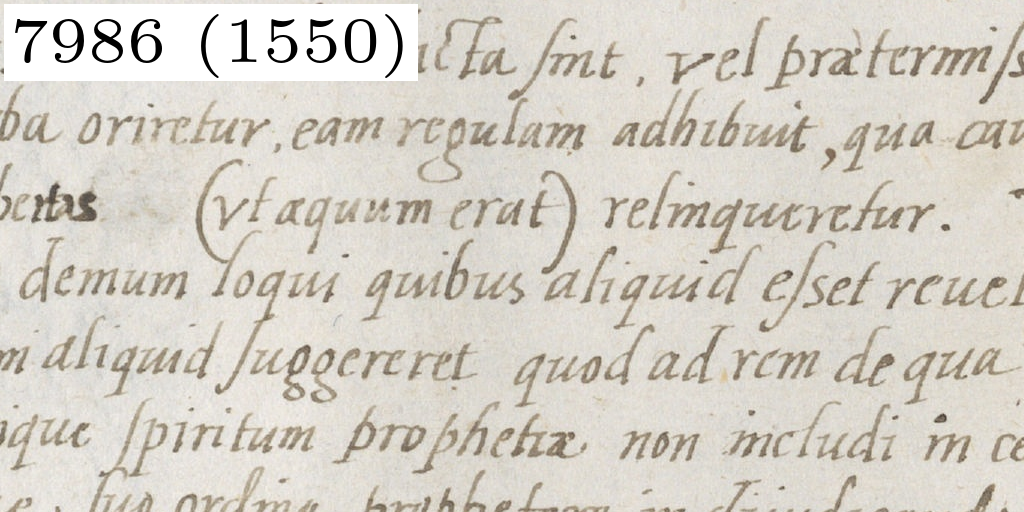}
}

\vspace{1mm}

\fcolorbox{chocolate2267451}{white}{
\includegraphics[width=0.9\linewidth]{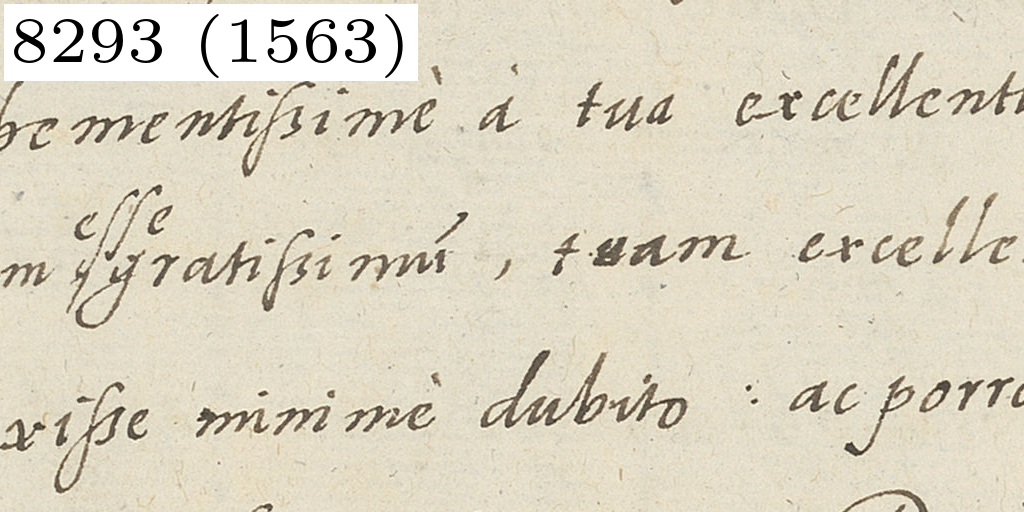}
}

\vspace{1mm}

\fcolorbox{chocolate2267451}{white}{
\includegraphics[width=0.9\linewidth]{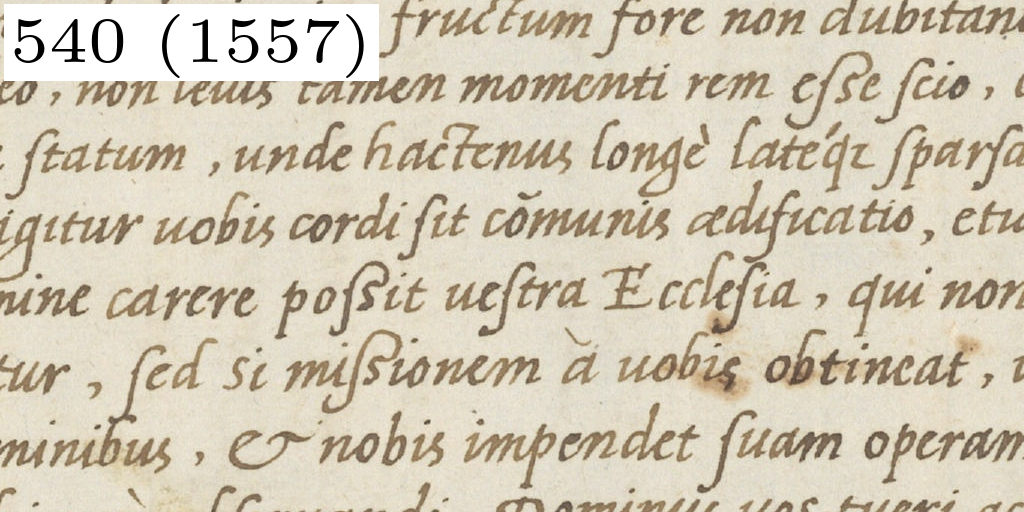}
}
\caption{ }
\end{subfigure}

\caption{\ac{WR} Visualization - Query: Blue, Green/Red are samples from the same/a different writer, \texttt{Writer (Year)}: (a) and (b) are handwritings of Heinrich Bullinger (37 years apart). (c) represents a non-successful retrieval, for which only the first page is correct. It also shows the variety of handwriting styles of \bull, in particular the stylistic change of Bullinger over the years.}\label{fig:wr-viz}
\end{figure}

%% file: figs/cer/freq/plot_1_cer_vs_length.tex
\begin{tikzpicture}

\definecolor{chocolate2267451}{RGB}{226,74,51}
\definecolor{dimgray85}{RGB}{85,85,85}
\definecolor{gainsboro229}{RGB}{229,229,229}

\begin{axis}[
axis background/.style={fill=gainsboro229},
axis line style={white},
tick align=outside,
tick pos=left,
width=0.48\linewidth,
height=0.4\linewidth,
x grid style={white},
xlabel=\textcolor{dimgray85}{Line length},
xmajorgrids,
xmin=-3.7965, xmax=150.8635,
xtick style={color=dimgray85},
y grid style={white},
ylabel=\textcolor{dimgray85}{CER},
ymajorgrids,
ymin=0.0524879298627929, ymax=0.63879561130157,
ytick style={color=dimgray85},
yticklabel={\pgfmathparse{\tick*100}\pgfmathprintnumber{\pgfmathresult}}
]
\addplot [semithick, chocolate2267451]
table {%
3.2415 0.669518398268398
5.725 0.295183723151349
8.2085 0.259041000521935
10.6915 0.214109838032861
13.175 0.213510102844267
15.6585 0.199541868445444
18.1415 0.15595070012908
20.625 0.123012309941427
23.1085 0.135005459675831
25.5915 0.156657711282434
28.075 0.123056539709565
30.5585 0.13198500069922
33.0415 0.115204541082319
35.525 0.105078929737344
38.0085 0.0952425296642012
40.4915 0.091964949362005
42.975 0.0859681615235602
45.4585 0.0822497567552966
47.9415 0.0841649324627102
50.425 0.0801848068360475
52.9085 0.0810929495542092
55.3915 0.0808087027497517
57.875 0.0788399366931753
60.3585 0.077935336117157
62.8415 0.0791323792807205
65.325 0.0886093857733703
67.8085 0.0797503437074537
70.2915 0.0767759615163657
72.775 0.0781310914047594
75.2585 0.0841751243763347
77.7415 0.0753944832976504
80.225 0.0839583013044588
82.7085 0.0751605701422321
85.1915 0.0792009041908636
87.675 0.0854521696524084
90.1585 0.0823965056882642
92.6415 0.0862777687344102
95.125 0.0921456350563059
97.6085 0.0970550194206622
100.0915 0.089786207324045
102.575 0.104475221546924
105.0585 0.10531304090495
107.5415 0.0866483987258396
110.025 0.0859125829828297
112.5085 0.0894075740911196
114.9915 0.117804999545142
117.475 0.0757816814268347
119.9585 0.118739902194804
122.4415 0.14641320947458
124.925 0.0924426773069574
127.4085 0.112903225806452
129.8915 0.10677857081974
132.375 0.110294117647059
134.8585 0.106060921716657
137.3415 0.101328464062665
139.825 0.12874752828259
142.3085 0.135135135135135
144.7915 0.104931972789116
};
\end{axis}

\end{tikzpicture}

%% file: figs/cer/freq/plot_2_mean_cer_per_writer.tex
\begin{tikzpicture}

\definecolor{chocolate2267451}{RGB}{226,74,51}
\definecolor{dimgray85}{RGB}{85,85,85}
\definecolor{gainsboro229}{RGB}{229,229,229}

\begin{axis}[
axis background/.style={fill=gainsboro229},
axis line style={white},
tick align=outside,
tick pos=left,
width=0.48\linewidth,
height=0.4\linewidth,
x grid style={white},
xlabel=\textcolor{dimgray85}{Writer Rank},
xmajorgrids,
xmin=0, xmax=143.5,
xtick style={color=dimgray85},
y grid style={white},
ylabel=\textcolor{dimgray85}{CER},
ymajorgrids,
ymin=0.0524879298627929, ymax=0.63879561130157,
ytick style={color=dimgray85},
yticklabel={\pgfmathparse{\tick*100}\pgfmathprintnumber{\pgfmathresult}}
]
\addplot [semithick, chocolate2267451]
table {%
0 0.0764970075226502
1 0.0775298904614578
2 0.0679793149648312
3 0.223549633127334
4 0.0672815123447394
5 0.0521674042917727
6 0.118823909709842
7 0.0662314464984175
8 0.0756689941041691
9 0.0669703293076865
10 0.111985144013738
11 0.0970204258247917
12 0.0966167055935947
13 0.0502967025282989
14 0.191806088744299
15 0.0781616776300825
16 0.0896823325474519
17 0.0701935054991214
18 0.0389939310562338
19 0.0476862594554873
20 0.0452049473331904
21 0.0779001692694964
22 0.121050684261734
23 0.0479525665656514
24 0.0749014775046588
25 0.0597741206550148
26 0.148389745605213
27 0.0892975748056252
28 0.18138635525826
29 0.196974014613742
30 0.188269479494757
31 0.14248571214681
32 0.121153500274243
33 0.111732244624854
34 0.0493426977197462
35 0.10519384228875
36 0.0357489530098709
37 0.103329590364818
38 0.0705739748760011
39 0.0328852138048063
40 0.0717778320261291
41 0.0374657228236164
42 0.071915897728398
43 0.0762084352207238
44 0.148179107991917
45 0.2213631430065
46 0.130386349642581
47 0.0595280550730259
48 0.0581375972056567
49 0.0588298606028708
50 0.102058452805275
51 0.152207119589067
52 0.0715127624286677
53 0.0751093158722108
54 0.0431445014645163
55 0.117438378604894
56 0.0777252837393974
57 0.0472659827355512
58 0.0694527257275508
59 0.143074743025557
60 0.150945291460971
61 0.166618847497395
62 0.0603764244232694
63 0.0805985435192731
64 0.0639412444738996
65 0.332604632010126
66 0.0610552183307469
67 0.0386696648092909
68 0.112812614303836
69 0.046769966198933
70 0.201468857715938
71 0.127931012581246
72 0.125150348856215
73 0.0818912654946353
74 0.0627597008784694
75 0.0580949914733679
76 0.0712568675408881
77 0.171794525024085
78 0.122352063041527
79 0.0785783135823768
80 0.11530284843189
81 0.148855372027986
82 0.133248857788233
83 0.110903176046866
84 0.0872349461269113
85 0.0508176561840016
86 0.0761826807490589
87 0.158125140974949
88 0.245020696736382
89 0.0317166574798117
90 0.0726827432071366
91 0.0619533165764676
92 0.0616347471038426
93 0.0823957583520174
94 0.0594300579695252
95 0.0725108060377216
96 0.138097086863236
97 0.117775317285459
98 0.0480323458722482
99 0.431205614211402
100 0.0906178299558676
101 0.0544350637854208
102 0.168347289159733
103 0.137059892700851
104 0.0415765586552093
105 0.0350550600149524
106 0.133018316250189
107 0.346978096645148
108 0.184860992138764
109 0.126064013428582
110 0.0935551067860452
111 0.21562341369574
112 0.234558702039489
113 0.0819547142645467
114 0.0655632686710487
115 0.0568187421639567
116 0.0518517892834441
117 0.0722034801194891
118 0.112806841171382
119 0.100551855457435
120 0.177026916931167
121 0.0540459060496581
122 0.277437805174585
123 0.137407331301198
124 0.179207107964585
125 0.0968773326061778
126 0.0969548900129002
127 0.124031581699932
128 0.0894602460063878
129 0.0861919103317364
130 0.0772666703094289
131 0.222471786371135
132 0.0463056213233975
133 0.123027533875747
134 0.0397307734384528
135 0.300819719022625
136 0.0775935350661803
137 0.10690232199648
138 0.114084388069539
139 0.0425001854797652
140 0.138841988785082
141 0.065781519897719
142 0.14240276095279
143 0.0403614216561168
};
\draw (axis cs:99,0.331205614211402) ++(5pt,5pt) node[
  scale=0.75,
  anchor=south west,
  text=black,
  rotate=0.0
]{8296};
\draw (axis cs:107,0.246978096645148) ++(5pt,5pt) node[
  scale=0.75,
  anchor=south west,
  text=black,
  rotate=0.0
]{8415};
\draw (axis cs:65,0.232604632010126) ++(5pt,5pt) node[
  scale=0.75,
  anchor=south west,
  text=black,
  rotate=0.0
]{152};
\end{axis}

\end{tikzpicture}

%% file: figs/cer/nonfreq/plot_1_cer_vs_length.tex
\begin{tikzpicture}

\definecolor{chocolate2267451}{RGB}{226,74,51}
\definecolor{dimgray85}{RGB}{85,85,85}
\definecolor{gainsboro229}{RGB}{229,229,229}
\definecolor{nonfreqblue}{RGB}{31,119,180}

\begin{axis}[
axis background/.style={fill=gainsboro229},
axis line style={white},
tick align=outside,
tick pos=left,
x grid style={white},
width=0.48\linewidth,
height=0.4\linewidth,
xlabel=\textcolor{dimgray85}{Line length},
xmajorgrids,
xmin=-3.5225, xmax=150.86355,
xtick style={color=dimgray85},
y grid style={white},
ylabel=\textcolor{dimgray85}{CER},
ymajorgrids,
ymin=0.0539438561033139, ymax=0.715647420306494,
ytick style={color=dimgray85},
yticklabel={\pgfmathparse{\tick*100}\pgfmathprintnumber{\pgfmathresult}}
]
\addplot [semithick, nonfreqblue]
table {%
3.1085 0.656011351909185
5.325 0.360419276214255
7.5415 0.261182599508227
9.7585 0.324476632294991
11.975 0.297575596991052
14.1915 0.283498857662937
16.4085 0.24506118383885
18.625 0.193279574582504
20.8415 0.193357264062178
23.0585 0.162684769476787
25.275 0.14546525555056
27.4915 0.155396961208411
29.7085 0.138704931644923
31.925 0.145877264563401
34.1415 0.134802423965753
36.3585 0.130333515135365
38.575 0.126863255482424
40.7915 0.123432564918744
43.0085 0.120104854378789
45.225 0.111830631523005
47.4415 0.10649126065469
49.6585 0.106468825091113
51.875 0.101343648134375
54.0915 0.100646284513681
56.3085 0.0946395941690893
58.525 0.0926393634168068
60.7415 0.0895810240441647
62.9585 0.0859762794741476
65.175 0.0818880018157604
67.3915 0.0848983364055592
69.6085 0.0817962146407917
71.825 0.0815111473748939
74.0415 0.0811260511476751
76.2585 0.0796583671574488
78.475 0.0908896131428922
80.6915 0.0874745789952482
82.9085 0.0897308252896282
85.125 0.0926847297918549
87.3415 0.0974430832306783
89.5585 0.0928371511967559
91.775 0.104864857871572
93.9915 0.12025818491986
96.2085 0.107366431857371
98.425 0.113765382620048
100.6415 0.117235407823334
102.8585 0.116661056505226
105.075 0.126911348570559
107.2915 0.123311436134585
109.5085 0.123849752093568
111.725 0.112114896376657
113.9415 0.134047647328903
116.1585 0.15087964143607
118.375 0.122256068900908
120.5915 0.12724004712304
122.8085 0.155952813568932
125.025 0.181905684144296
127.2415 0.228855721393035
129.4585 0.18801652892562
131.675 0.145438762626263
};
\end{axis}

\end{tikzpicture}

%% file: figs/cer/nonfreq/plot_2_mean_cer_per_writer.tex
\begin{tikzpicture}

\definecolor{chocolate2267451}{RGB}{226,74,51}
\definecolor{dimgray85}{RGB}{85,85,85}
\definecolor{gainsboro229}{RGB}{229,229,229}
\definecolor{nonfreqblue}{RGB}{31,119,180}

\begin{axis}[
axis background/.style={fill=gainsboro229},
axis line style={white},
tick align=outside,
tick pos=left,
x grid style={white},
width=0.48\linewidth,
height=0.4\linewidth,
xlabel=\textcolor{dimgray85}{Writer Rank},
xmajorgrids,
xmin=0, xmax=651.5,
xtick style={color=dimgray85},
y grid style={white},
ylabel=\textcolor{dimgray85}{CER},
ymajorgrids,
ymin=-0.0492841662940614, ymax=1.54282333046431,
ytick style={color=dimgray85},
yticklabel={\pgfmathparse{\tick*100}\pgfmathprintnumber{\pgfmathresult}}
]
\addplot [semithick, nonfreqblue]
table {%
0 0.0556733933580899
1 0.0520243201270819
2 0.107487347139114
3 0.0803355999363147
4 0.118248522146562
5 0.0569427515069991
6 0.0881610232643674
7 0.148873510377138
8 0.12514627488881
9 0.125219834880602
10 0.130534350146531
11 0.105608549259299
12 0.0433954414964506
13 0.106394868848595
14 0.179881987935457
15 0.0922220405447353
16 0.142456877019757
17 0.175125247783755
18 0.116026337152863
19 0.108634934038849
20 0.112181341833827
21 0.162919579254305
22 0.105509975530292
23 0.0745344207350817
24 0.111962607268866
25 0.220735495092872
26 0.0897407453956386
27 0.133202686747428
28 0.0908509211594066
29 0.129168417757312
30 0.103082602660035
31 0.105220430909068
32 0.180270185712363
33 0.161685891177684
34 0.0674932348955842
35 0.0937721756824236
36 0.0949201201137355
37 0.0768546169310429
38 0.107007122570633
39 0.245147460217678
40 0.0619449613884326
41 0.0552879377077647
42 0.170749727943266
43 0.0595714630495206
44 0.0881106990843425
45 0.124606187754746
46 0.109255766482338
47 0.121500192591329
48 0.0856390752542808
49 0.109354395210089
50 0.0725689609919029
51 0.171889819864334
52 0.0703528600630981
53 0.0730902980351851
54 0.10798738762746
55 0.137351481506271
56 0.107874794660761
57 0.0912079500269026
58 0.0826775851613096
59 0.135203780095961
60 0.0831914516331534
61 0.0844560130194807
62 0.0874966391590165
63 0.0673465448611742
64 0.0479384852898248
65 0.079174850315027
66 0.138149697050823
67 0.143834941118086
68 0.137308240889971
69 0.0911148908319379
70 0.0751016320678982
71 0.105962771012485
72 0.0754501105829786
73 0.0525472869458806
74 0.0846250953108609
75 0.044681636591434
76 0.378521441469558
77 0.0810997905986786
78 0.1259678675755
79 0.175845232167053
80 0.0923013425608748
81 0.154114731020586
82 0.131380897726505
83 0.129702901004702
84 0.12418021410061
85 0.121791127989073
86 0.123589648187783
87 0.0689090055782247
88 0.0808248713643089
89 0.125109469512014
90 0.0686843950086972
91 0.0545204477254595
92 0.141159723041508
93 0.119533917654414
94 0.099327337503454
95 0.0849046331915892
96 0.105345384830759
97 0.0585586774481631
98 0.0831898687498372
99 0.0353832021241702
100 0.205937702209667
101 0.185867398377804
102 0.06466191743662
103 0.0737745424235477
104 0.0664419732801235
105 0.208031329829167
106 0.0648169094002879
107 0.0878987964075792
108 0.0628961074204379
109 0.0510165797943785
110 0.0920557457383716
111 0.0399645243302191
112 0.0686568546424375
113 0.124594323026154
114 0.087435636312879
115 0.0699680683600844
116 0.10119593536633
117 0.0473226719060034
118 0.0958863410775226
119 0.0819285113457563
120 0.0687860394384509
121 0.0717999588735678
122 0.140421620119035
123 0.125209355677049
124 0.102021292901537
125 0.0691585295730709
126 0.15212378744434
127 0.0866101235333737
128 0.207966098444814
129 0.202848250121322
130 0.138357273155956
131 0.276988180900454
132 0.0705012745868658
133 0.114021869121196
134 0.120742775663603
135 0.0800029409139818
136 0.0667296817784969
137 0.1349643437156
138 0.0737228592567482
139 0.154888994461701
140 0.151112090460523
141 0.167014586153819
142 0.077589620141484
143 0.0566120562308273
144 0.102458866968767
145 0.163770082094281
146 0.0631119934101325
147 0.0569332198503231
148 0.15496820333742
149 0.0903290353531537
150 0.0545550524560067
151 0.0765066491588232
152 0.207479703914603
153 0.0955490515236697
154 0.124015127855507
155 0.0733750764689438
156 0.203325311382124
157 0.078121950329404
158 0.0435916791714025
159 0.112034417274369
160 0.347841909366314
161 0.1163704666491
162 0.0409754743433071
163 0.100062508621617
164 0.0737720757312095
165 0.109050105134778
166 0.109009964053908
167 0.0696011561407134
168 0.228789554575456
169 0.111666618948129
170 0.178789956207102
171 0.0688417681045173
172 0.130949845296496
173 0.0914571522950329
174 0.169526431946764
175 0.0986292924118855
176 0.0499759260034511
177 0.0779904856231423
178 0.117343382998125
179 0.068544995178617
180 0.0860942234107739
181 0.0381078173321833
182 0.0527765307542029
183 0.148868682147932
184 0.0487125566753151
185 0.14029027087396
186 0.0685013408475664
187 0.209635467561719
188 0.163342417322811
189 0.084652008352449
190 0.0978961996564942
191 0.0648082135602178
192 0.0961988400622775
193 0.0576114401280899
194 0.0695794309272237
195 0.140585658823137
196 0.157533946821705
197 0.144933866135607
198 0.0918327247394479
199 0.160882478167903
200 0.241397399012189
201 0.0649956364315229
202 0.11963887217947
203 0.109680374204505
204 0.0936245180070072
205 0.0670862000581473
206 0.0305730593236878
207 0.0767720595222895
208 0.122528009824796
209 0.0858096932272342
210 0.0604415372416958
211 0.108952745917024
212 0.0717262536953803
213 0.0814807619569977
214 0.071149971572004
215 0.0508323071663387
216 0.132738101410706
217 0.151934388903777
218 0.0627304795024647
219 0.179171137045474
220 0.139940569762591
221 0.0577706509694366
222 0.114932040679713
223 0.074382899119099
224 0.0395668027760503
225 0.138831744265646
226 0.120010033097874
227 0.0882909442896122
228 0.0379712916615876
229 0.0779689278509887
230 0.0438922979514161
231 0.0430843600402573
232 0.0795571806305945
233 0.0594328188858579
234 0.316874187978595
235 0.0869962983335215
236 0.0818210744803547
237 0.0794993590900934
238 0.0608071109239311
239 0.0524070764610347
240 0.0518795664689271
241 0.0605648183165345
242 0.0714817945599645
243 0.12982960342213
244 0.084889090832232
245 0.0698171584637033
246 0.145803941835718
247 0.0469895019473502
248 0.162527405704951
249 0.35668325242869
250 0.0917563072246084
251 0.108760939326801
252 0.0638363428090668
253 0.505713093826778
254 0.133551639517168
255 0.105499586838335
256 0.0485833537541462
257 0.238163751704105
258 0.0254976465700504
259 0.0927322189539315
260 0.0380416990293847
261 0.191241287849663
262 0.0951031730009056
263 0.0989008343564412
264 0.195427547597703
265 0.120093432784921
266 0.158901458735117
267 0.126513038922044
268 0.12068909346384
269 0.0806520292845734
270 0.152320445496546
271 0.0854054273316974
272 0.0547953059422992
273 0.0887037721793338
274 0.137144853355295
275 0.0791227637385712
276 0.0583970132343536
277 0.103754442437675
278 0.0911415289188796
279 0.0532015233741954
280 0.0691293033956065
281 0.0798149705128986
282 0.116957418309431
283 0.0643792276320627
284 0.0874141355038321
285 0.0485208662245498
286 0.109471577886828
287 0.0950990872955634
288 0.217296171067042
289 0.107717504220644
290 0.0803943553983376
291 0.0382114010804478
292 0.094448925390416
293 0.157351578953778
294 0.0214782304519187
295 0.136045936758838
296 0.0503365407758054
297 0.0839260860989446
298 0.0623144178158791
299 0.0325519801437346
300 0.094918759084939
301 0.0932589248976559
302 0.0944228224770972
303 0.0636012514002511
304 0.154904113427389
305 0.228930996189956
306 0.0670369030619775
307 0.142979422725372
308 0.125807650847767
309 0.359845317127634
310 0.105488731995255
311 0.0905835814475668
312 0.0628574013585784
313 0.102334690981883
314 0.0653279014354463
315 0.0530495008511795
316 0.123834580204429
317 0.0391308300775933
318 0.0675820577596711
319 0.0496291944536629
320 0.145385942294116
321 0.101325676439202
322 0.10703820135984
323 0.099010319968629
324 0.350107142776408
325 0.0639323334191065
326 0.0833518561277507
327 0.0411169113754775
328 0.112307238073109
329 0.135148515651637
330 0.0639326676104873
331 0.114596582598631
332 0.0643424512716021
333 0.126895629953942
334 0.0721947488555676
335 0.0764302580065648
336 0.146390548440374
337 0.0781245118997885
338 0.0877252107051172
339 0.145290482840016
340 0.0819448567424529
341 0.129691095885682
342 0.0721459496965334
343 0.162949357779677
344 0.0890623598294615
345 0.120435422945092
346 0.101136077074287
347 0.0923884412824426
348 0.201351766095328
349 0.0342928442101794
350 0.102324708018193
351 0.0581590080965183
352 0.0342184992807661
353 0.0985624410802699
354 0.188740234734339
355 0.10261269782296
356 0.173167117413952
357 0.0414752460411084
358 0.0828973494954169
359 0.14563379365449
360 0.197081698237356
361 0.11652991205519
362 0.0752515092771
363 0.113011638922562
364 0.067056242108974
365 0.129529264638747
366 0.323444462455934
367 0.130785284169089
368 0.137028506161173
369 0.0875665705799451
370 0.186746169834172
371 0.0791059571516112
372 0.0510201159275909
373 0.0576002485362702
374 0.282589018864459
375 0.11574933881132
376 0.0634741759337773
377 0.0721649939910182
378 0.189583693803171
379 0.181611867764961
380 0.19154899130818
381 0.144140937319651
382 0.0400956172680384
383 0.168720658139002
384 0.108895497345734
385 0.0938899914139751
386 0.0450114651349932
387 0.0928215266543823
388 0.0712329763680182
389 0.0934299264495492
390 0.132302542332505
391 0.0564895073939046
392 0.179986276524714
393 0.0613511495932903
394 0.210937809971443
395 0.242290892325591
396 0.151961507779163
397 0.116950046683189
398 0.185556190556236
399 0.275204907863236
400 0.123594717373304
401 0.100288075261311
402 0.371503487602253
403 0.154498500086472
404 0.089385903246054
405 1.18576162368662
406 0.129917835549392
407 0.165395517839697
408 0.0461698605723679
409 0.0600322043923233
410 0.171101408505514
411 0.0840539487819141
412 0.0997941018162823
413 0.140974526386007
414 0.209227424974823
415 0.0617396539860666
416 0.182536315266356
417 0.0528328107446449
418 0.126476740312972
419 0.0870587955487199
420 0.0445673075174468
421 0.0635773673393621
422 0.0887520606567356
423 0.137528369231091
424 0.074046003381708
425 0.305310439160476
426 0.143951851337713
427 0.0749118180649204
428 0.212570797220339
429 0.166222564960517
430 0.0818392972066937
431 0.104890786294936
432 0.124029866074116
433 0.0605387242568251
434 0.366199942629702
435 0.0763932186452261
436 0.0654028695806261
437 0.143644894207631
438 0.082306891927113
439 0.112477186426592
440 0.113842542522011
441 0.0406604840450686
442 0.069816166561576
443 0.0472574850828529
444 0.0696734227373356
445 0.0574550206324803
446 0.0476284104815219
447 0.130515529144998
448 0.230014895734849
449 0.0606959885135333
450 0.0997912742040395
451 0.109940918207956
452 0.187162265439358
453 0.0409301526039977
454 0.230687292869287
455 0.130571193809063
456 0.09009309980021
457 0.15236389869596
458 0.0680959346118974
459 0.154309985066719
460 0.0997553810207252
461 0.92905783967776
462 0.0495521476000775
463 0.0652713891128782
464 0.0565165951868011
465 0.0572918266351188
466 0.0647559111756326
467 0.156317267737036
468 0.123670323303453
469 0.120553248289194
470 0.0511611911735295
471 0.0729285650290212
472 0.0560610593951961
473 0.0616778417400703
474 0.109895051307109
475 0.142113938485786
476 0.086266684218088
477 0.160391732019749
478 0.0665258463995588
479 0.0554880513722815
480 0.0414358847159393
481 0.0995216394996135
482 0.13598884272291
483 0.0438972457755217
484 0.0855891398996413
485 0.130225881831384
486 0.100444100085827
487 0.0898087543240103
488 0.228404844421556
489 0.235927459252612
490 0.0800053858136034
491 0.0874373082081013
492 0.0583948308215342
493 0.121117945642237
494 0.367846949229142
495 0.105089936906308
496 0.21542471767383
497 0.0537437007426125
498 0.180130887187746
499 0.0578510764423838
500 0.108934342618225
501 0.0956264589882937
502 0.171308372846815
503 0.042637403192862
504 0.0427978115828206
505 0.110918644539075
506 0.0648129194981053
507 0.0289295632751101
508 0.0308567296784107
509 0.116297188286173
510 0.179993862247308
511 0.082158960607382
512 0.0595021286945428
513 0.0279450939269274
514 0.102661284219677
515 0.0217381835050623
516 0.0652423547866226
517 0.10361893231376
518 0.13252695449256
519 0.0964014187092171
520 0.221866910426707
521 0.097830336862072
522 0.0560608679576513
523 0.215637344255947
524 0.107111063025197
525 0.0441194273350755
526 0.0307185971521469
527 0.0487594533561672
528 0.308483941769311
529 0.0993478712190109
530 0.04797730194937
531 0.189420684396478
532 0.746405388132152
533 0.0432907586248876
534 0.0964624556571234
535 0.230120256858124
536 0.0858007223076251
537 0.18511669642103
538 0.0986330557698384
539 0.111835883733978
540 0.279957544870398
541 0.0399212689403226
542 0.262798389520419
543 0.10744458434087
544 0.139597090398861
545 0.0388058124396058
546 0.0759801668426867
547 0.169455631237027
548 0.0555090435025928
549 0.16745403752596
550 0.0780470201178273
551 0.0778447909083816
552 0.50450059476101
553 0.111478069706241
554 0.123322400737107
555 0.0803396626643036
556 0.0570579400476459
557 0.108530188718755
558 0.0664465040344485
559 0.178316955400055
560 0.121961574071599
561 0.274720936820154
562 0.219372028164272
563 0.0530457291170588
564 0.248674896018939
565 0.042967399003129
566 0.150034199000282
567 0.170010491431519
568 0.236844593521176
569 0.285824941151031
570 0.0851259906008728
571 0.0686423142305832
572 0.240313955374913
573 0.0935168794270039
574 0.132148369524156
575 0.049662590507449
576 0.0538361192865924
577 0.0740074720604511
578 0.122017257958184
579 0.0532050779727308
580 0.0999774141309715
581 0.0658725777581599
582 0.249810237839923
583 0.103458927105299
584 0.0420690852932424
585 0.101874291257084
586 0.247574138049278
587 0.069569858215917
588 0.0675417902594911
589 0.0872017842300086
590 0.0782847042641997
591 0.206606572494539
592 0.223925578064734
593 0.111193518925022
594 0.124339536031006
595 0.104921508552147
596 0.130045025439959
597 0.091003307566365
598 0.0826868173651851
599 0.0468540486000629
600 0.178215198866453
601 0.150704071742099
602 0.109183575350605
603 0.219091470057889
604 0.390748773619181
605 0.143336989501506
606 0.0825371089351536
607 0.0875220285845458
608 0.138997763328781
609 0.0330608438588771
610 0.130545623264457
611 0.0738561394054484
612 0.0460534377247595
613 0.00887049083382614
614 0.062138813880845
615 0.149946183936582
616 0.147003426735936
617 0.0793214065256063
618 0.160897930453988
619 0.0493794367892119
620 0.0313395138882333
621 0.108632036477305
622 0.374279114188914
623 0.158174726017036
624 0.0780442948036291
625 0.224749886916852
626 0.132288546090053
627 0.241167307256491
628 0.0846769027754579
629 0.201347601541121
630 0.13669215901239
631 0.0805832291376756
632 0.0553081870840666
633 0.135941484394243
634 0.123434213785471
635 0.138316520914505
636 0.106789560685879
637 0.131540584836195
638 0.110210347079087
639 0.079688659048538
640 0.30523888377735
641 0.211364408289684
642 0.046224912535928
643 0.107198555698397
644 0.0566927231893775
645 0.027352509824882
646 0.0753512120603545
647 0.167894604190721
648 0.2728392030292
649 0.0671167263390927
650 0.105744824637323
651 0.0954780136231749
};
\draw (axis cs:405,1.08576162368662) ++(5pt,5pt) node[
  scale=0.75,
  anchor=south west,
  text=black,
  rotate=0.0
]{8457};
\draw (axis cs:461,0.82905783967776) ++(5pt,5pt) node[
  scale=0.75,
  anchor=south west,
  text=black,
  rotate=0.0
]{1455};
\draw (axis cs:532,0.646405388132152) ++(5pt,5pt) node[
  scale=0.75,
  anchor=south west,
  text=black,
  rotate=0.0
]{8514};
\end{axis}
\end{tikzpicture}

%% file: sections/6_conclusion.tex
\section{Conclusion}\label{sec:conclusion}

In this work, we introduced \bull, a new benchmark dataset for \ac{HTR} and \ac{WR}. The dataset comprises 20,898 pages and about 500k text lines written by 796 distinct writers, making it a substantial and valuable resource for historical document analysis. We evaluated current state-of-the-art approaches on both tasks, achieving a \ac{CER} of 9.1\% for \ac{HTR} and a \ac{mAP} of approximately 78.3 \% for \ac{WR}. These results demonstrate that \bull \ is a challenging benchmark dataset for the community and highlight potential room for further research on this historically relevant collection.
For future work, large-scale \ac{HTR} models could be trained from scratch or pre-trained using self-supervised learning strategies. Writer adaptation also represents a promising direction, as performance varies across writers. However, intra-writer variability introduces an additional challenge, as handwriting styles can evolve over time (e.g., changes observed in Bullinger’s handwriting throughout different periods).
Regarding \ac{WR}, the temporal dimension might be studied more carefully. In particular, embedding disentanglement could be explored by modeling representations as $\boldsymbol{z} = \boldsymbol{z}_\mathrm{w} + \boldsymbol{z}_\mathrm{t}$, where $\boldsymbol{z}_\mathrm{w}$ captures writer-specific characteristics that remain invariant over time, and $\boldsymbol{z}_\mathrm{t}$ encodes temporal information. Such an approach could first be examined on a smaller, potentially text-dependent scale, for example by focusing on specific words or individual text lines.

{\small
\subsubsection*{Acknowledgements} This work has been supported by the Hasler Foundation, Switzerland, and the Swiss National Science Foundation (SNSF) under project grant no. 10002280.
 We thank the contributors of the Bullinger Digital project, in particular Tobias Hodel, Anna Janka, Raphael Müller, Peter Rechsteiner, David Selim Schoch, Raphael Schwitter, Christian Sieber, Phillip Ströbel, Martin Volk, and Jonas Widmer for their contributions to the dataset and ground truth creation.}

%% file: bib.bib
@inproceedings{bullinger-dataset2023,
  author       = {Anna Scius{-}Bertrand and
                  Phillip Str{\"{o}}bel and
                  Martin Volk and
                  Tobias Hodel and
                  Andreas Fischer},
  title        = {The Bullinger Dataset: {A} Writer Adaptation Challenge},
  booktitle    = {Document Analysis and Recognition - {ICDAR} 2023 - 17th International
                  Conference, San Jos{\'{e}}, CA, USA, August 21-26, 2023, Proceedings,
                  Part {I}},
  pages        = {397--410},
  year         = {2023},
}

@inproceedings{jungo2023,
  author       = {Michael Jungo and
                  Lars V{\"{o}}gtlin and
                  Atefeh Fakhari and
                  Nathan Wegmann and
                  Rolf Ingold and
                  Andreas Fischer and
                  Anna Scius{-}Bertrand},
  title        = {Impact of the ground truth quality for handwriting recognition},
  booktitle    = {Proceedings of the 12th International Symposium on Information and
                  Communication Technology, {SOICT} 2023, Vietnam, December
                  7-8, 2023},
  pages        = {135--140},
  year         = {2023},
}

@article{stroebel2024multilingual,
  title        = {Multilingual Workflows in Bullinger Digital: Data Curation for Latin and Early New High German},
  author       = {Str{\"o}bel, Phillip Benjamin and Fischer, Lukas and M{\"u}ller, Raphael and Scheurer, Patricia and Schroffenegger, Bernard and Suter, Benjamin and Volk, Martin},
  journal      = {Journal of Open Humanities Data},
  volume       = {10},
  number       = {12},
  pages        = {1--13},
  year         = {2024},
}

@inproceedings{Peer2025,
author = {Peer, Marco and Scius-Bertrand, Anna and Fischer, Andreas},
booktitle = {2nd International Workshop on Computer Vision Systems for Document Analysis and Recognition (VisionDocs@ICCV2025)},
pages = {1–10},
title = {CTC Transcription Alignment of the Bullinger Letters: Automatic Improvement of Annotation Quality},
year = {2025}
}

@article{transkribus,
  title={Transforming scholarship in the archives through handwritten text recognition: Transkribus as a case study},
  author={Muehlberger, Guenter and Seaward, Louise and Terras, Melissa and Ares Oliveira, Sofia and Bosch, Vicente and Bryan, Maximilian and Colutto, Sebastian and D{\'e}jean, Herv{\'e} and Diem, Markus and Fiel, Stefan and others},
  journal={Journal of documentation},
  volume={75},
  number={5},
  pages={954--976},
  year={2019},
}

@INPROCEEDINGS{PyLaia,
  author={Puigcerver, Joan},
  booktitle={2017 14th IAPR International Conference on Document Analysis and Recognition (ICDAR)},
  title={Are Multidimensional Recurrent Layers Really Necessary for Handwritten Text Recognition?},
  year={2017},
  number={},
  pages={67-72},
}

@inbook{Gatos2004,
  title = {An Adaptive Binarization Technique for Low Quality Historical Documents},
  booktitle = {Document Analysis Systems VI},
  author = {Gatos,  Basilios and Pratikakis,  Ioannis and Perantonis,  Stavros J.},
  year = {2004},
  pages = {102–113}
}

@inproceedings{icdar19,
  author    = {Vincent Christlein and
               Anguelos Nicolaou and
               Mathias Seuret and
               Dominique Stutzmann and
               Andreas Maier},
  title     = {{ICDAR} 2019 Competition on Image Retrieval for Historical Handwritten
               Documents},
  booktitle = {2019 International Conference on Document Analysis and Recognition,
               {ICDAR} 2019, Sydney, Australia, September 20-25, 2019},
  pages     = {1505--1509},
  year      = {2019}
}

@inproceedings{historical-wi,
  author    = {Stefan Fiel and
               Florian Kleber and
               Markus Diem and
               Vincent Christlein and
               Georgios Louloudis and
               Stamatopoulos Nikos and
               Basilis Gatos},
  title     = {{ICDAR2017} Competition on Historical Document Writer Identification
               (Historical-WI)},
  booktitle = {14th {IAPR} International Conference on Document Analysis and Recognition,
               {ICDAR} 2017, Kyoto, Japan, November 9-15, 2017},
  pages     = {1377--1382},
  year      = {2017}
}

@misc{anyscript2026,
  title        = {AnyScript: Long-Term Handwriting Author Identification Challenge},
  author       = {{Computer Vision Center}, Universitat Aut\`onoma de Barcelona},
  year         = {2026},
  howpublished = {\url{https://cvc-dag.github.io/anyscript/}},
  note         = {Dataset and ICDAR 2026 competition for temporal handwriting author identification},
  organization = {Computer Vision Center (CVC) and Universitat Aut\`onoma de Barcelona}
}

@inproceedings{cvl,
  author    = {Florian Kleber and
               Stefan Fiel and
               Markus Diem and
               Robert Sablatnig},
  title     = {{CVL-DataBase}: An Off-Line Database for Writer Retrieval, Writer Identification
               and Word Spotting},
  booktitle = {12th International Conference on Document Analysis and Recognition,
               {ICDAR} 2013, Washington, DC, USA, August 25-28, 2013},
  pages     = {560--564},
  year      = {2013},
}

@article{marti_iam-database_2002,
	title = {The {IAM}-database: an {English} sentence database for offline handwriting recognition},
	volume = {5},
	issn = {1433-2833, 1433-2825},
	shorttitle = {The {IAM}-database},
	number = {1},
	urldate = {2021-05-08},
	journal = {International Journal on Document Analysis and Recognition},
	author = {Marti, U.-V. and Bunke, H.},
	year = {2002},
	pages = {39--46},
}

@article{Mauelshagen2010,
  title = {Heinrich Bullinger (1504-1575): Leben - Denken - Wirkung},
  ISSN = {0254-4407},
  journal = {Zwingliana},
  publisher = {Zwingliana},
  author = {Mauelshagen,  Franz},
  year = {2010},
  month = jan,
  pages = {89–106}
}

@book{HBBW_Reihe,
  editor       = {{Institut f{\"u}r Schweizerische Reformationsgeschichte (IRG)}},
  title        = {Heinrich Bullinger Werke. Briefwechsel},
  publisher    = {Theologischer Verlag Z{\"u}rich (TVZ)},
  address      = {Z{\"u}rich},
  year         = {1973--},
  note         = {Kritische Edition der Korrespondenz Heinrich Bullingers; mit Angaben zu Autographen, fremden H{\"a}nden, Archivsignaturen und editorischem Apparat}
}

@inproceedings{volk-etal-2022-nunc,
  title     = {Nunc profana tractemus. Detecting Code-Switching in a Large Corpus of 16th Century Letters},
  author    = {Volk, Martin and Fischer, Lukas and Scheurer, Patricia and Schroffenegger, Bernard Silvan and Schwitter, Raphael and Str{\"o}bel, Phillip and Suter, Benjamin},
  booktitle = {Proceedings of the Thirteenth Language Resources and Evaluation Conference},
  month     = jun,
  year      = {2022},
  address   = {Marseille, France},
  pages     = {2901--2908}
}

@inproceedings{lavrenko2004holistic,
  author    = {Lavrenko, Victor and Rath, Toni M. and Manmatha, Raghavan},
  title     = {Holistic Word Recognition for Handwritten Historical Documents},
  booktitle = {Proceedings of the First International Workshop on Document Image Analysis for Libraries (DIAL)},
  year      = {2004},
  pages     = {278--287},
  publisher = {IEEE},
}

@article{fischer2012lexiconfree,
  author  = {Fischer, Andreas and Keller, Andreas and Frinken, Volker and Bunke, Horst},
  title   = {Lexicon-Free Handwritten Word Spotting Using Character HMMs},
  journal = {Pattern Recognition Letters},
  volume  = {33},
  number  = {7},
  pages   = {934--942},
  year    = {2012},
}

@inproceedings{Retsinas2022,
  title = {Best Practices for a Handwritten Text Recognition System},
  booktitle = {Document Analysis Systems},
  author = {Retsinas,  George and Sfikas,  Giorgos and Gatos,  Basilis and Nikou,  Christophoros},
  year = {2022},
  pages = {247–259}
}

@inproceedings{deSousaNeto2020,
  booktitle = {DocEng ’20},
  title = {HTR-Flor++: A Handwritten Text Recognition System Based on a Pipeline of Optical and Language Models},
  author = {de Sousa Neto,  Arthur Flor and Bezerra,  Byron Leite Dantas and Toselli,  Alejandro Héctor and Lima,  Estanislau Baptista},
  year = {2020},
  pages = {1–4},
}

@BOOK{Fischer2020-al,
  title     = "Handwritten historical document analysis, recognition, and
               retrieval - state of the art and future trends",
  author    = "Fischer, Andreas and Liwicki, Marcus and Ingold, Rolf",
  editor    = "Fischer, Andreas and Liwicki, Marcus and Ingold, Rolf",
  series    = "Series In Machine Perception And Artificial Intelligence",
  month     =  nov,
  year      =  2020,
  address   = "Singapore, Singapore",
  language  = "en"
}

@inproceedings{Graves2006,
  title = {Connectionist temporal classification: labelling unsegmented sequence data with recurrent neural networks},
  booktitle = {Proceedings of the 23rd international conference on Machine learning  - ICML ’06},
  author = {Graves,  Alex and Fernández,  Santiago and Gomez,  Faustino and Schmidhuber,  J\"{u}rgen},
  year = {2006},
  pages = {369–376},
  collection = {ICML ’06}
}

@article{Hochreiter1997,
  title = {Long Short-Term Memory},
  volume = {9},
  number = {8},
  journal = {Neural Computation},
  author = {Hochreiter,  Sepp and Schmidhuber,  J\"{u}rgen},
  year = {1997},
  month = nov,
  pages = {1735–1780}
}

@inproceedings{trocr,
  author       = {Minghao Li and
                  Tengchao Lv and
                  Jingye Chen and
                  Lei Cui and
                  Yijuan Lu and
                  Dinei A. F. Flor{\^{e}}ncio and
                  Cha Zhang and
                  Zhoujun Li and
                  Furu Wei},

  title        = {TrOCR: Transformer-Based Optical Character Recognition with Pre-trained
                  Models},
  booktitle    = {Thirty-Seventh {AAAI} Conference on Artificial Intelligence, {AAAI}
                  2023, Thirty-Fifth Conference on Innovative Applications of Artificial
                  Intelligence, {IAAI} 2023, Thirteenth Symposium on Educational Advances
                  in Artificial Intelligence, {EAAI} 2023, Washington, DC, USA, February
                  7-14, 2023},
  pages        = {13094--13102},
  year         = {2023},
}

@inproceedings{fischer-etal-2022-machine,
    title = "Machine Translation of 16{T}h Century Letters from {L}atin to {G}erman",
    author = "Fischer, Lukas  and
      Scheurer, Patricia  and
      Schwitter, Raphael  and
      Volk, Martin",
    booktitle = "Proceedings of the Second Workshop on Language Technologies for Historical and Ancient Languages",
    month = jun,
    year = "2022",
    address = "Marseille, France",
    pages = "43--50",
}

@inproceedings{unsupervised_icdar17,
  author    = {Vincent Christlein and
               Martin Gropp and
               Stefan Fiel and
               Andreas K. Maier},
  title     = {Unsupervised Feature Learning for Writer Identification and Writer
               Retrieval},
  booktitle = {14th {IAPR} International Conference on Document Analysis and Recognition,
               {ICDAR} 2017, Kyoto, Japan, November 9-15, 2017},
  pages     = {991--997},
  year      = {2017}
}

@inproceedings{peer_netrvlad,
  author       = {Marco Peer and
                  Florian Kleber and
                  Robert Sablatnig},
  title        = {Towards Writer Retrieval for Historical Datasets},
  booktitle    = {Document Analysis and Recognition - {ICDAR} 2023 - 17th International
                  Conference, San Jos{\'{e}}, CA, USA, August 21-26, 2023},
  pages        = {411--427},
  year         = {2023},
}

@inproceedings{raven,
  author       = {Tim Raven and
                  Arthur Matei and
                  Gernot A. Fink},

  title        = {Self-supervised Vision Transformers for Writer Retrieval},
  booktitle    = {Document Analysis and Recognition - {ICDAR} 2024 - 18th International
                  Conference, Athens, Greece, August 30 - September 4, 2024, Proceedings,
                  Part {II}},
  pages        = {380--396},
  year         = {2024},
}

@inproceedings{peer2024,
  author       = {Marco Peer and
                  Florian Kleber and
                  Robert Sablatnig},
  title        = {{SAGHOG:} Self-supervised Autoencoder for Generating {HOG} Features
                  for Writer Retrieval},
  booktitle    = {Document Analysis and Recognition - {ICDAR} 2024 - 18th International
                  Conference, Athens, Greece, August 30 - September 4, 2024},
  pages        = {121--138},
  year         = {2024},
}

@misc{norhand,
  url = {https://zenodo.org/doi/10.5281/zenodo.10255839},
  author = {Beyer,  Yngvil and Solberg,  Per Erik},
  language = {no},
  title = {NorHand v3 / Dataset for Handwritten Text Recognition in Norwegian},
  publisher = {Zenodo},
  year = {2023},
  copyright = {Creative Commons Attribution 4.0 International}
}

@inproceedings{peer2022,
  author       = {Marco Peer and
                  Florian Kleber and
                  Robert Sablatnig},
  title        = {Self-supervised Vision Transformers with Data Augmentation Strategies
                  Using Morphological Operations for Writer Retrieval},
  booktitle    = {Frontiers in Handwriting Recognition - 18th International Conference,
                  {ICFHR} 2022, Hyderabad, India, December 4-7, 2022, Proceedings},
  pages        = {122--136},
  year         = {2022}
}

@inbook{Strbel2023,
  title = {The Adaptability of a Transformer-Based OCR Model for Historical Documents},
  booktitle = {Document Analysis and Recognition – ICDAR 2023 Workshops},
  author = {Str\"{o}bel,  Phillip Benjamin and Hodel,  Tobias and Boente,  Walter and Volk,  Martin},
  year = {2023},
  pages = {34–48}
}

@inbook{Spoto2022,
  title = {Improving Handwriting Recognition for Historical Documents Using Synthetic Text Lines},
  booktitle = {Intertwining Graphonomics with Human Movements},
  author = {Spoto,  Martin and Wolf,  Beat and Fischer,  Andreas and Scius-Bertrand,  Anna},
  year = {2022},
  pages = {61–75}
}
